\documentclass[letterpaper, 10 pt, journal, twoside]{ieeetran}  

\IEEEoverridecommandlockouts                              

\usepackage{cite}
\usepackage{graphicx}
\graphicspath{{figs/}}
\usepackage{amsmath}
\usepackage{amssymb}
\usepackage{algorithm}
\usepackage{algpseudocode}
\usepackage{bm}
\usepackage{physics}
\usepackage{siunitx}
\usepackage{subcaption}
\usepackage{times}
\newcommand{\equref}[1]{\eqref{#1}}
\newcommand{\figref}[1]{Fig.~\ref{#1}}
\newcommand{\tabref}[1]{Table~\ref{#1}}
\newcommand{\secref}[1]{Sec.\ref{#1}}

\allowdisplaybreaks

\newif\ifrevise
\revisefalse 
\newcommand{\revise}[1]{\textcolor{\ifrevise blue\else black\fi}{#1}}

\def\paperlanguage{}

\newcommand{\switchlanguage}[2]{%
  \ifx\paperlanguage\empty%
  #1%
  \else%
  #2%
  \fi%
}

\title{
Anti-Gravity Walking by a Flying Humanoid \\Robot via Thrust-Rate Input Whole-Body \\Model Predictive Control
}

\author{Kazuki Sugihara$^{1}$ and Kei Okada$^{1}$ 
  \thanks{Manuscript received: May, 19, 2026; Revised July, 29, 2026; Accepted September, 6, 2026.}
  \thanks{This paper was recommended for publication by Editor Giuseppe Loianno upon evaluation of the Associate Editor and Reviewers' comments.}
  \thanks{This work was supported by JSPS KAKENHI Grant Number JP25KJ0776.}
  \thanks{$^{1}$Kazuki Sugihara and Kei Okada are with Department of Mechano-Infomatics, The University of Tokyo, 7-3-1 Hongo, Bunkyo-ku, Tokyo 113-8656, Japan.
    {\tt\footnotesize sugihara@jsk.imi.i.u-tokyo.ac.jp}}
  \thanks{Digital Object Identifier (DOI): see top of this page.}
}

\begin{document}

\maketitle

\markboth{IEEE Robotics and Automation Letters. Preprint Version. Accepted September, 2026}
         {Sugihara \MakeLowercase{\textit{et al.}}: Anti-Gravity Walking via Thrust-Rate Input WB MPC}

\begin{abstract}
\switchlanguage{
Flying humanoid\revise{s} are expected to perform tasks in diverse environments, while their existing locomotion is \revise{mainly} limited to aerial flight and ground walking.
The capability to move in complex three-dimensional space can greatly expand their application range.
For such walking motion on ceilings and similar anti-gravity environments, whole-body MPC is effective.
However, the discontinuous changes in dynamic structure accompanying contact switching during walking can induce thrust spikes, resulting in control instability.
Therefore, in this work, we propose and implement a real-time whole-body MPC framework \revise{for anti-gravity bipedal walking}.
First, we formulate whole-body MPC using the time derivative of thrust, namely thrust-rate, as the control input.
This formulation guarantees continuity of the thrust trajectory during contact switching while preserving the sparse structure of the optimal control problem for fast computation.
Second, we address the lack of natural support forces in anti-gravity environments.
We introduce lower bound\revise{s} on the foot-normal component of the contact force, and smoothly transfer \revise{them} during the double-support phase.
Finally, we implement the proposed framework and demonstrate anti-gravity walking by a flying humanoid through simulation \revise{and a hardware experiment}.
To the best of our knowledge, this is the first demonstration of multi-contact whole-body MPC for a transformable aerial robot and walking by a flying humanoid beyond the ground.
}{
飛行ヒューマノイドは多様な環境での活動が期待されるが、既存のロコモーションは空中や地上に限られている.
高所作業において環境接触により機体を安定化させつつ移動する能力は, その応用範囲を飛躍的に拡大させる.
このような, 天井などの抗重力環境での歩行には動的変化に対応する全身MPCが有効である.
しかし歩行時の接触状態の切り替えに伴う不連続な力学的構造の変化が, 推力が急激に変化するスパイクを誘発し, 実機への適用の大きな障壁となっていた.
本論文では、飛行ヒューマノイドによる天井での抗重力歩行を実現可能にするリアルタイム全身MPCフレームワークを提案・実装する.
第一に, 推力の時間微分(Thrust-rate)を制御入力とする全身MPCの定式化により, 接触切り替え時の推力軌道の連続性の保証しつつ, 問題のスパース性を維持することで高速な求解を可能にした.
第二に, 重力による支持力が得られない環境特有の課題に対し, 足裏法線方向の最小接触力制約を導入し, さらに両脚支持期にこれを滑らかに遷移させる荷重移動戦略を組み込んだ. 
そして, 我々はこれらに基づく全身MPC frameworkを実装し, 物理シミュレータを用いた評価により, 飛行ヒューマノイドによる抗重力歩行が可能なことを実証した.
To the best of our knowledge, 変形飛行ロボットによる多点接触全身MPCおよび飛行ヒューマノイドによる地上以外での歩行を実現した初の例である.
}
\end{abstract}

\section{Introduction}
\switchlanguage{
\IEEEPARstart{F}{lying} humanoid\revise{s} with walking and flight capabilities have recently attracted attention as versatile platforms for various tasks in diverse environments \cite{kim2021bipedal,anzai2021design,gorbani2025ironcub,xu2025system,sugihara2023design}.
Although they have achieved seamless walking and flying \cite{kim2021bipedal,anzai2021design}, their locomotion capabilities are still limited to aerial flight and walking \revise{on nearly flat ground}.
In tasks near elevated structures, environmental contact can help stabilize the body and reduce the reliance on continuous hovering.
Therefore, enabling flying humanoids to walk in anti-gravity environments such as ceilings would expand their application range, and open a new direction for locomotion in aerial robots and humanoids.
}{
近年, 飛行ヒューマノイドプラットフォームの研究が盛んで, 脚式移動と空中飛行の能力を統合したプラットフォームとして、多様な環境での複雑なタスク遂行が期待されている \cite{kim2021bipedal,anzai2021design,gorbani2025ironcub,xu2025system,sugihara2023design}.
プロペラと脚機構を協調させて歩行と飛行をシームレスに行うLEONARDO\cite{kim2021bipedal}や, 全身の自由度を有する飛行ヒューマノイド\cite{anzai2021design}が開発されたが, これらのロコモーションは依然として空中飛行や地上歩行に限られている.
また, 飛行ロボットの高所での物理的な接触を伴うタスクにおいて, 継続的なホバリングは姿勢の安定性の観点で不利であり, 機体を環境に接触させることで安定化させるアプローチも提案されている \cite{nishio2023perchingarm}.
以上より, 飛行ヒューマノイドがこのような天井などの抗重力環境で歩行する能力は, 飛行ロボット, ヒューマノイドによるロコモーションの新たな地平を切り開くとともに, その応用範囲を飛躍的に拡大させる.
}

\begin{figure}
    \centering
    \includegraphics[width=1.0\columnwidth]{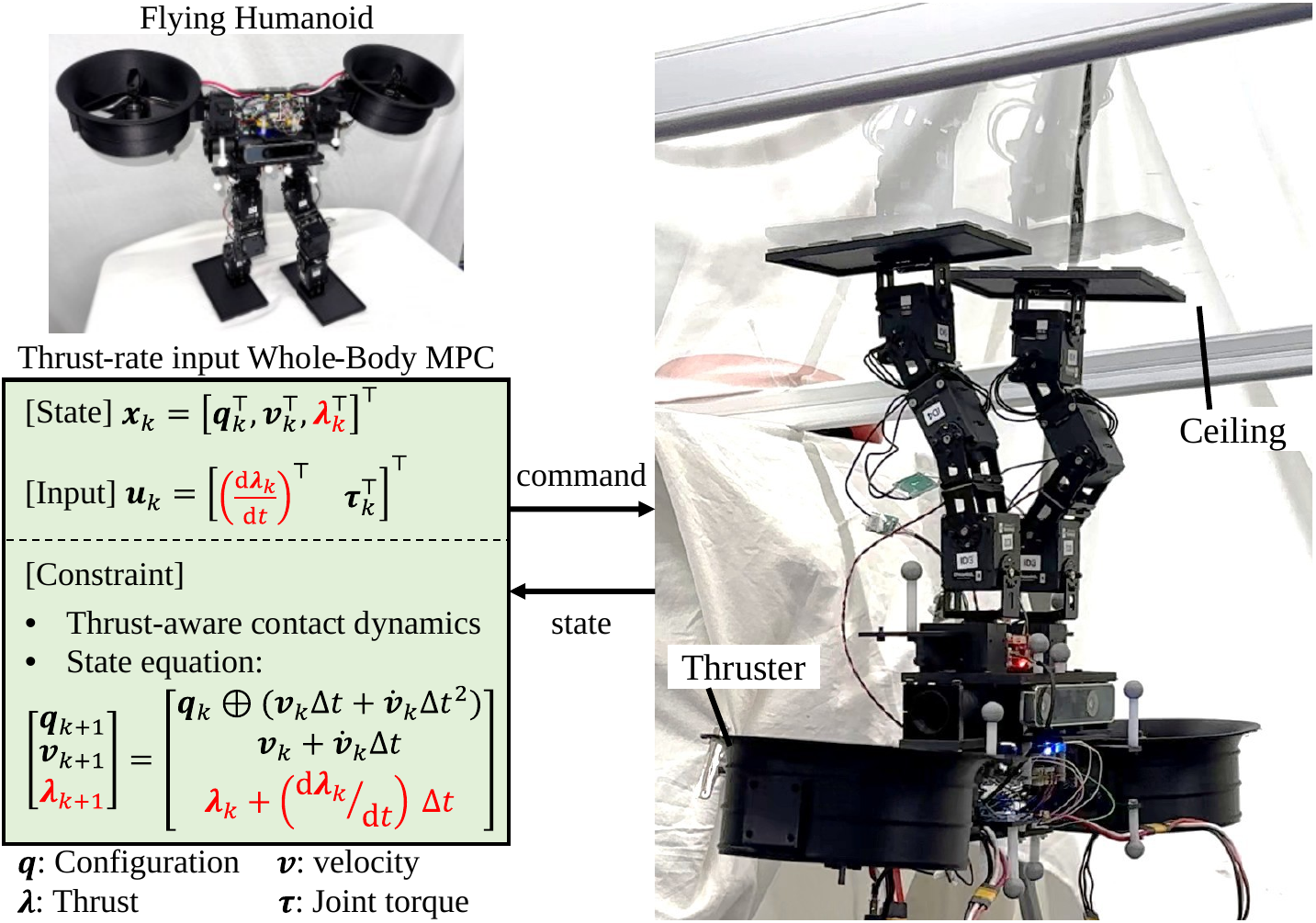}
    \caption{\revise{Overview of the proposed thrust-rate input whole-body MPC} and Anti-gravity walking by a flying humanoid.}
    \vspace{-2mm}
    \label{figure:overview}
\end{figure}

\switchlanguage{
As an approach to handle such multi-contact and whole-body dynamics, whole-body model predictive control (whole-body MPC) has been applied to legged robots \cite{khazoom2024tailoring,boxfddp}.
In anti-gravity walking, where natural support from gravity is not available, the robot must be continuously pressed against the environment by thrust to ensure sufficient friction.
However, when the dynamic structure changes with foot contact switching, a standard MPC formulation using thrust as the input may require step-like changes in thrust to \revise{satisfy} dynamic equilibrium.
Actual hardware cannot easily track such spike-like commands, which has been a major challenge for real implementation.
Furthermore, naively introducing input rate constraints to ensure thrust continuity would destroy the sparse structure of the optimal control problem (OCP) and make real-time computation difficult.
Therefore, realizing walking in anti-gravity environments requires an MPC formulation that simultaneously handles thrust generation as an aerial robot, contact switching as a legged robot, and whole-body dynamics with multi-contact, while ensuring continuity of thrust.
In this work, we propose a whole-body MPC that treats thrust itself as a state and the time derivative of thrust, namely thrust-rate as control input.
With this formulation, since thrust is calculated by integrating the thrust-rate, continuity of the thrust can be ensured without adding difference constraints between control inputs.
Furthermore, to maintain foot contact in anti-gravity environments, we introduce lower bound\revise{s} on the foot-normal component of the contact force into the contact wrench cone (CWC) \revise{condition} and incorporate a load transfer strategy that smoothly change these bounds during double-support phase.
By combining these elements, we construct a whole-body MPC framework that enables anti-gravity walking on ceilings by a flying humanoid robot \revise{as shown in \figref{figure:overview}}.
}{
このような多点接触と全身の動力学を扱うアプローチとして、全身モデル予測制御(Whole-body MPC)が脚式ロボット分野で用いられ始めている\cite{khazoom2024tailoring,mastalli2023inverse}.
重力による自然な支持が得られない抗重力環境での歩行では, 十分な摩擦力を確保するため, ロータ推力によって常にロボットを環境へ押し付ける必要がある.
しかし, 足先の接触状態の切替えに伴って力学的な構造が変化する際, 一般的なMPCの定式化では動的平衡を保つためにロータ推力に対してステップ状の急変を要求しうる.
実際のハードウェアではこのようなスパイク的な指令値への追従は難しく, これが実機への適用の大きな障壁となっていた.
さらに, 推力の連続性を担保するために安易な入力差分制約(Rate constraints)を導入すれば、最適制御問題におけるスパース構造が破壊され, リアルタイムでの求解が困難になるという問題もあった.
そのため, 抗重力環境での歩行を実現するためには, 飛行ロボットとしての推力生成, 脚式ロボットとしての接触切替え, そして多点接触全身動力学を同時に扱いながら, ロータ推力の時間連続性を保つMPC定式化が必要である.
本研究ではこの問題に対し, 推力そのものを状態として扱い, 推力の時間微分を入力とする全身MPCを提案する.
この定式化では, 推力をthrust-rateの積分で算出するため, 入力間の差分制約を追加することなく, 発揮推力の連続性を保つことができる.
さらに, 抗重力環境における足裏接触を維持するため, 接触レンチ錐の制約に最小法線力を導入し, 両脚支持期に左右足間でこの下限を連続的に遷移させる荷重移動戦略を組み込む.
これらの要素を組み合わせることで, 飛行ヒューマノイドによる天井での抗重力歩行を実現可能にする全身MPCフレームワークを構築する.
}

\switchlanguage{
The key contributions of this paper are as follows:
\begin{itemize}
    \item We propose a thrust-rate input whole-body MPC framework that mathematically guarantees the continuity of the thrust trajectory during contact switching while preserving the sparse structure of the problem.
    \item To address challenges specific to walking in anti-gravity environments, we introduce lower bounds on the foot-normal component of the contact force and transfer strategy that smoothly changes these bounds during double-support phase.
    \item We integrate the proposed methods into a real-time control system and demonstrate anti-gravity walking by a flying humanoid through simulation \revise{and real machine}.
\end{itemize}
}{
本論文の主な貢献は以下の通りである.
\begin{itemize}
    \item Thrust-rateを入力にする全身MPCフレームワークを提案し, 問題のスパース性を維持したまま、接触状態の遷移に伴う推力軌道の連続性を数理的に保証した。
    \item 抗重力環境に特有の課題に対処するため, 両脚支持期における足裏法線方向の最小接触力制約を歩行フェーズに応じて滑らかに遷移させる荷重移動戦略を導入した.
    \item 提案手法を実時間制御システムとして統合し, 物理シミュレータ環境下で飛行ヒューマノイドによる抗重力歩行が可能なことを初めて実証した.
\end{itemize}
}

\section{Related Work}
\label{sec:related_work}
\subsection{Thrust\revise{er}\revise{-Equipped} Legged Robots}
\switchlanguage{
Legged robots with thrusters have been developed mainly for extending ground motion and for achieving aerial flight.
\revise{Platforms such as} LEONARDO, \revise{SPIDAR, and Husky integrated legged and aerial locomotion in a single body \cite{kim2021bipedal, zhao2023spidar, wang2025dynamic}}.
\revise{During ground contact, thrust was used to support the body, for example to reduce joint loads \cite{zhao2023spidar} or to traverse a large gap \cite{liu2018jet}.}
\revise{Thrust was also used to actively shape the contact forces so that otherwise infeasible contact conditions were satisfied.
Harpy stabilized the frontal dynamics of a biped to satisfy gait feasibility conditions \cite{dangol2021control}, and the thrusters of Husky kept the contact forces inside the friction cone on slopes of up to \SI{45}{\degree} \cite{salagame2024quadrupedal}.}
\revise{For flight, iRonCub3, a jet-powered humanoid, demonstrated a liftoff \cite{gorbani2025ironcub}, and} configuration optimization during flight \revise{was} proposed for a flying humanoid with redundant DoF \cite{zhang2025flight}.

In existing thruster-assisted walking, gravity presses the feet onto a floor or a slope, and thrust only reduces the required contact force or steers it inside the friction cone.
On ceilings and walls, whose surface normal differs greatly from the gravity direction, the normal force itself must be generated by thrust, and walking has not been addressed.
}{
推力装置を搭載した脚ロボットは推力を利用した地上動作の拡張と空中飛行の研究を中心に発展してきた.
LEONARDOは, プロペラ推力を利用した二足歩行と飛行を実現し, 推力支援によって地上での運動性能を拡張できることを示した\cite{kim2021bipedal}.
ロータ分散型4脚ロボットでは推力を利用して脚部の負荷軽減をしながら地上歩行が行われている \cite{zhao2023spidar}.
Jet-HR1は足部に搭載したダクトファン推力を用いて, 大きな段差やgapをまたぐ動作を実現した \cite{liu2018jet}.
また, 冗長自由度を有する飛行ヒューマノイドの空中姿勢最適化も提案されている \cite{zhang2025flight}.

これらの脚ロボットによる既存のロコモーションは依然として地上歩行と空中飛行に限られている.
脚は主に地上での支持・歩行に用いられ, ロータ推力は飛行, 跳躍, 姿勢安定化, または関節負荷の軽減を担う補助的な役割に留まる.
天井や壁面のように重力方向と接触面法線が大きく異なる環境では, 足裏接触を維持するための法線力が重力によって自然には得られないため, このような環境での歩行は扱われてこなかった.
これらに対し, 本研究は推力を用いて足裏での接触力を生み出すことで飛行ヒューマノイドのロコモーションを地上・空中から抗重力環境での歩行へと広げることを目指す.
}

\subsection{Environmental Contact with Multirotor Platform}
\switchlanguage{
In aerial manipulation, aerial robots equipped with grippers or arms generate\revise{d} interaction forces with the environment using thrust \cite{ollero2022aerialmanipulationreview, bodie2019omnidirectionalmanipulation}.
In addition, an aerial manipulator utilized contact \revise{for} stable manipulation \cite{nishio2023perchingarm}.

However, these approaches typically design joint motion and thrust allocation separately.
\revise{As a result,} joint torques, thrusts, whole-body dynamics, and multi-contact constraints \revise{are not treated in a unified manner}.
In anti-gravity walking, thrust must continuously generate the contact forces required to keep each support foot attached to the surface while satisfying contact wrench constraints.
Therefore, in this work, \revise{the joint motion and the thrust allocation are coupled through the whole-body dynamics, and determined within a single OCP.}
}{
ロータ推力を環境への接触力として能動的に生成するアプローチは, 飛行ロボット分野で複数の文脈で発展してきた.
空中マニピュレーションでは, 飛行機体に把持機構やアームを搭載し, 推力により環境へ作業力を加える手法が提案されている \cite{ollero2022aerialmanipulationreview, kim2013aerial, bodie2019omnidirectionalmanipulation}.
また, 環境への接触を利用してホバリング単体より安定な作業姿勢を得る飛行マニピュレータが提案された \cite{nishio2023perchingarm}.
これらの研究では, 接触は主に作業対象へ力を加えるため, または機体を安定化するための拘束として利用される.

しかしながらこれらの研究では, 関節軌道とロータ推力配分が分離して設計されていて, 関節とロータが全身動力学と多点接触を含む単一の最適制御問題として統合的に扱われていない.
抗重力歩行は, 各支持脚で接触レンチ錐制約を満たす法線力と摩擦力をロータ推力で継続的に生成し続ける必要がある.
重力による自然な支持が得られないため, 関節軌道と推力配分を分離して設計するアプローチでは, 接触切替に伴う力学的構造の変化と推力連続性を扱いきれない.
そこで本研究は, 関節トルクとロータ推力を全身動力学および多点接触制約のもとで同一の最適制御問題内で決定する.
}

\subsection{Thrust-Aware Whole-Body MPC}
\switchlanguage{
Whole-body MPC handles whole-body dynamics, contact, and input constraints simultaneously.
\revise{It is thus effective for legged locomotion, and hardware deployments have recently been realized \cite{khazoom2024tailoring}}.
\revise{Contact-implicit MPC further removes the need for pre-planned contact modes \cite{kim2025contactimplicit}, and learning-based controllers have also achieved agile humanoid locomotion \cite{radosavovic2024real}}.
For aerial robots, MPC ha\revise{s} treated thrust as an input together with multi-link dynamics \cite{shi2019dynamics}, and contact \cite{martisaumell2023fullbodytorquelevelnonlinearmodel}.
\revise{Thrust can thus be added to} whole-body MPC as an \revise{extra} input.
However, \revise{this} does not guarantee thrust continuity, resulting in a difficulty for real implementation.
\revise{In the above approaches like RL as well, the smoothness of the commanded input is shaped by a penalty rather than by the formulation.}
For smooth input trajectory generation, \revise{a related formulation included the wrench in the state and used} wrench-rate as the input \revise{together} with constraints \cite{brunner2020trajectorytracking}.

\revise{In this work, we treat thrust as a state and its time derivative as the control input, without adding constraints that couple consecutive inputs, so that the sparse structure of the OCP is preserved.
The continuity of thrust is therefore guaranteed by the state transition itself, rather than encouraged by a penalty.}
}{
脚式ロボットの全身MPCは, 全身動力学, 接触拘束, 入力制約を同時に扱えるため, 接触切替えを伴う動的ロコモーションに有効である.
近年は解精度と計算時間のトレードオフを調整しながら高速な全身MPCを実現した\cite{khazoom2024tailoring}.
飛行ロボットに対するMPCでは, 推力を入力に含め, 多リンク動力学とともに扱う試み\cite{shi2019dynamics}や, 空中マニピュレータで接触を扱う例\cite{martisaumell2023fullbodytorquelevelnonlinearmodel}がある
以上を組み合わせてロータ推力を制御入力として全身MPCに加えることが考えられるが, 推力の連続性が担保できない.
これに対し, wrench rateを入力にして制約を追加することが行われた \cite{brunner2020trajectorytracking}.
そこで本研究では, 全身MPCにおいてロータ推力を状態として拡張し, 推力差分を制御入力とすることで, 推力連続性を状態遷移に埋め込む.
推力レート制約を直接追加しないため, シューティング法に基づくMPCのスパース構造を損なわず実時間での求解が可能になる.
また, 十分な接触力を稼ぎつつ接触力も同様になめらかに変化させるため, 足平での最小法線力制約を導入し, 両足支持期をつかって遷移させる戦略を提案する.
これにより, 多点接触とその切替え, 関節トルク制約, およびロータ推力制約を同一の全身MPC内で扱い, 抗重力歩行に必要な滑らかな推力軌道を生成する.
}

\section{Dynamics and Whole-Body MPC Formulation}
\label{sec:dynamics}

\subsection{Dynamics Model of Flying Humanoid Robot}

\subsubsection{Robot Dynamics}
\begin{figure}[t]
    \centering
    \includegraphics[width=1.0\columnwidth]{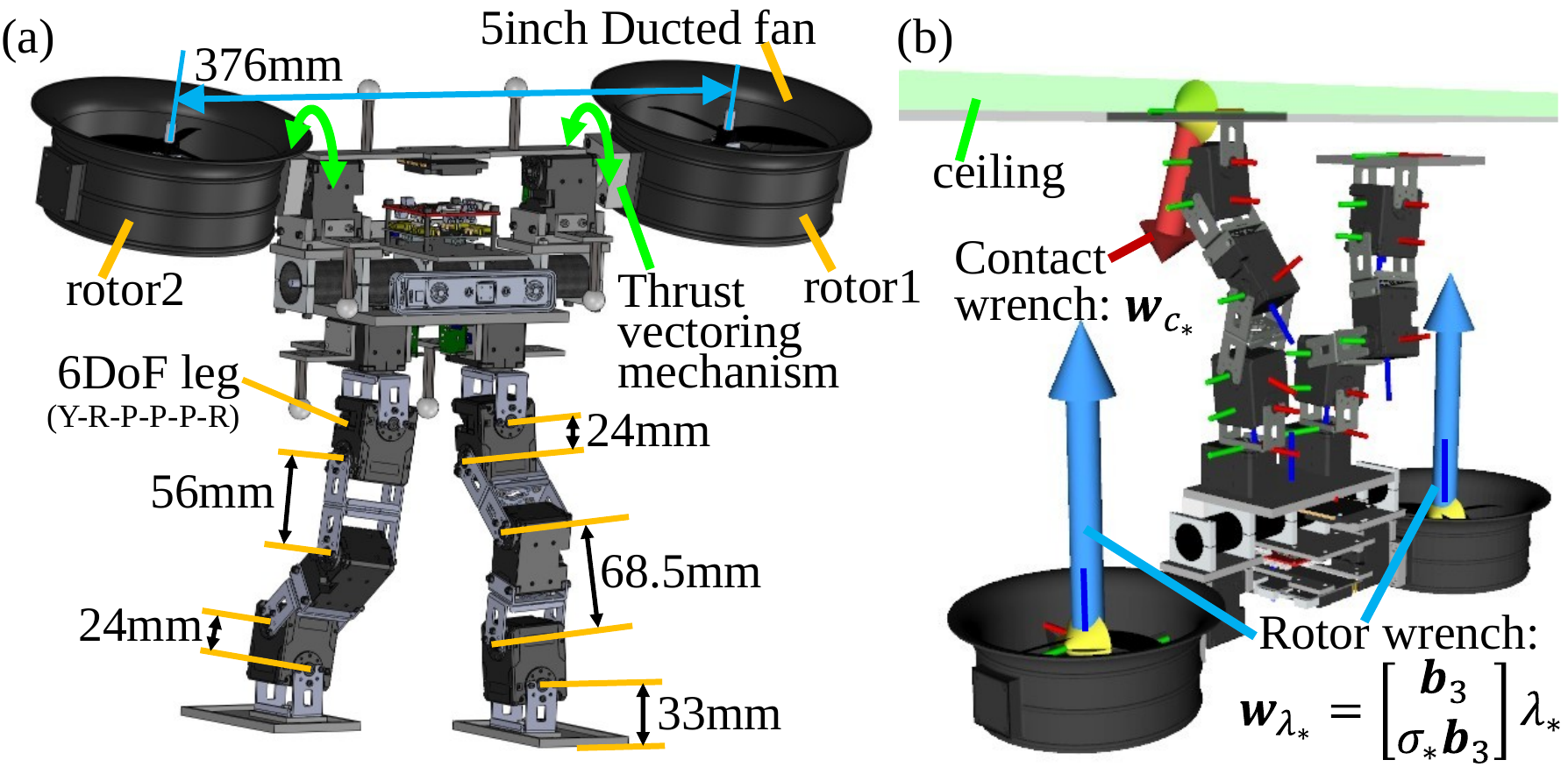}
    \caption{(a) Design of the flying humanoid which consists of two legs with six DoF and two vectorable thrusters. (b) Dynamics model of the flying humanoid walking on the ceiling.}
    \vspace{-2mm}
    \label{figure:dynamics}
\end{figure}

\switchlanguage{
The design \revise{and link length} of the flying humanoid considered in this work \revise{are} shown in \figref{figure:dynamics}(a).
The robot consists of two legs with six DoF \revise{with Yaw-Roll-Pitch-Pitch-Pitch-Roll configuration} and two thrusters.
Each thruster is equipped with \revise{1 DoF} thrust vectoring mechanism that change the thrust direction, enabling each rotor to rotate independently around the \(y\)-axis of the \revise{root} \revise{(}torso\revise{)} link.
This allows the robot to generate moments around each axis with a smaller number of thrusters.
Furthermore, by appropriately controlling these angles according to the inclination of the contact surfaces such as walls and ceilings, the rotor thrust can be directed toward the contact surface normal.
This is advantageous for efficiently generating the pressing force required for walking in anti-gravity environments.
}{
本研究で扱う飛行ヒューマノイドの設計を\figref{figure:dynamics}(a)に示す.
各脚6自由度からなり, 更に推力方向を可変にする推力偏向装置を搭載しており, ルートリンク\(y\)軸回りに各ロータを独立に回転させることができる.
これによって各軸周りの回転モーメントを少ないロータ数で生成可能である.
さらに, この関節角を壁面や天井といった傾斜に応じて適切に制御することで, ロータ推力を接触面法線方向に向けることができ, 抗重力環境での歩行に必要な押し付け力を効率的に生成できるという利点もある.
}

\switchlanguage{
The dynamics model of this flying humanoid is shown in \figref{figure:dynamics}(b), and the equations of motion are as follows,
}{
この飛行ヒューマノイドのdynamics modelは\figref{figure:dynamics}(b)に示すもので, 運動方程式は以下である.
}
\newcommand{\zaxis}{\bm{b}_3}
\newcommand{\transpose}{^\top}
\newcommand{\thrustwrench}[1]{\bm{w}_{\lambda_{#1}}}
\newcommand{\jacobian}[1]{\bm{J}_{#1}}
\newcommand{\Nc}{N_{\text{c}}}
\newcommand{\Nr}{N_{\text{r}}}
\newcommand{\Njoint}{N_{\text{joint}}}
\newcommand{\Nv}{N_{\text{v}}}
\begin{equation}
	\label{eq:robot_dynamics}
\begin{aligned}
\bm{M}(\bm{q})\dot{\bm{v}} + \bm{C}(\bm{q}, \bm{v})\bm{v} + \bm{g}(\bm{q}) =
\begin{bmatrix}
    \bm{0}_6\\
    \bm{\tau}
\end{bmatrix}
\\\quad+ \sum_{i=1}^{\Nr} \jacobian{\lambda_i}\transpose 
        \revise{
            \begin{bmatrix}
                \zaxis\\
                \sigma_i\zaxis
            \end{bmatrix}\lambda_i}
+ \sum_{i=1}^{\Nc} \jacobian{c_i}\transpose \bm{w}_{c_i}.
\end{aligned}
\end{equation}
\switchlanguage{
In \eqref{eq:robot_dynamics}, \(\bm{q} \in \mathcal{Q} := \mathbb{SE}(3) \times \mathbb{R}^{\Njoint}\) denotes robot configuration including the floating base and joint angles, and \(\bm{v} \in T_{\bm{q}}\mathcal{Q} \simeq \mathfrak{se}(3) \times \mathbb{R}^{\Njoint}\) denotes the generalized velocity in its tangent space.
Here, \(\Njoint\), \(\Nr\), and \(\Nc\) denote the number of joints, rotors, and contact points, respectively.
We treat \(\mathfrak{se}(3) \simeq \mathbb{R}^6\), then we write \(\bm{v} \in \mathbb{R}^{\Nv}\), where \(\Nv = 6+\Njoint\).
\(\bm{M}(\bm{q})\) is the inertia matrix, \(\bm{C}(\bm{q}, \bm{v})\bm{v}\) is the Coriolis and centrifugal term, \(\bm{g}(\bm{q})\) is the gravity term, \revise{and} \(\bm{\tau} \in \mathbb{R}^{\Njoint}\) is the joint torque.
\revise{In this work, we use rotors to generate thrust, so thrust in the direction of the \(z\)-axis of the rotor frame and drag moment proportional to thrust are generated.
\(\lambda_i, \sigma_i \in \mathbb{R}\) are defined as the thrust and drag moment coefficients (signed) of the \(i\)-th rotor, respectively, and \(\zaxis\) is defined as \(\begin{bmatrix}0 & 0 & 1\end{bmatrix}\transpose\).}
\(\bm{w}_{c_i} \in \mathbb{R}^6\) denotes the contact wrench acting at the \(i\)-th contact frame.
\(\bm{0}_{*}\) denotes a \(*\)-dimensional zero vector.
\(\jacobian{\lambda_i}\)\revise{,} \(\jacobian{c_i}\) \revise{\(\in \mathbb{R}^{6\times \Nv}\)}are the Jacobians of the \(i\)-th rotor frame and the \(i\)-th contact frame, respectively.
}{
\eqref{eq:robot_dynamics}において, \(\bm{q} \in \mathcal{Q} := \mathbb{SE}(3) \times \mathbb{R}^{\Njoint}\)は浮遊基底と関節角を含むロボットのコンフィギュレーション, \(\bm{v} \in T_{\bm{q}}\mathcal{Q} \simeq \mathfrak{se}(3) \times \mathbb{R}^{\Njoint}\)はその接空間における一般化速度を表す.
ただし\(\Njoint\)は関節数, \(\Nr\)はロータ数, \(\Nc\)は接触点数を表す.
ここでは\(\mathfrak{se}(3) \simeq \mathbb{R}^6\)とみなし, \(\Nv = 6+\Njoint\)を用いて, \(\bm{v} \in \mathbb{R}^{\Nv}\)と書く.
\(\bm{M}(\bm{q})\)は慣性行列, \(\bm{C}(\bm{q}, \bm{v})\bm{v}\)はコリオリ力と遠心力, \(\bm{g}(\bm{q})\)は重力項, \(\bm{\tau} \in \mathbb{R}^{\Njoint}\)は関節トルクを表す.
本研究では推力を生み出すロータに回転翼を用い, ロータ座標系\(z\)軸に沿った方向への推力とそれに比例する大きさの反トルクが発生するとする.
\(\lambda_i, \sigma_i \in \mathbb{R}\)はそれぞれ\(i\)番目ロータの発揮推力, 反トルク係数(符号付き), \(\zaxis = \begin{bmatrix}0 & 0 & 1\end{bmatrix}\transpose\)と定義する.
\(\bm{0}_{*}\)は\(*\)次元のゼロベクトルを表す.
\(\bm{w}_{c_i} \in \mathbb{R}^6\)は\(i\)番目接触点に作用する接触レンチを表す.
\(\jacobian{\lambda_i}\), \(\jacobian{c_i}\)はそれぞれ\(i\)番目のロータ座標系, 接触点のヤコビアンである.
}

\subsection{Whole-body MPC Formulation}
\switchlanguage{
The whole-body MPC is formulated as a discrete-time OCP as follows,
}{
全身MPCは以下の離散時間最適制御問題として表される.
}
\newcommand{\Nx}{N_{\text{x}}}
\newcommand{\Nu}{N_{\text{u}}}
\newcommand{\statetraj}{\text{X}}
\newcommand{\inputtraj}{\text{U}}
\begin{align}
 &\underset{\substack{\statetraj=\{\bm{x}_0 \cdots \bm{x}_N\}\\ \inputtraj=\{\bm{u}_0 \cdots \bm{u}_{N-1}\}}}{\text{minimize}} \hspace{2mm} && \sum_{k = 0}^{N-1} l_{k}(\bm{x}_k, \bm{u}_k) + l_N(\bm{x}_N), \label{eq:ocp_cost}\\
 &&&\bm{x}_{k+1} = \bm{f}\qty(\bm{x}_{k}, \bm{u}_k), \label{eq:ocp_state_eq}\\
 &&& \revise{\bm{c}}(\bm{x}_k, \bm{u}_k) \leq \bm{0}, \label{eq:ocp_ineq_con}\\
 &&& \bm{h}(\bm{x}_k, \bm{u}_k) = \bm{0}, \label{eq:ocp_eq_con}\\
 &&&\bm{x}_0 = \bm{x}(0). \label{eq:ocp_initial_variable}
\end{align}
\switchlanguage{
In \eqref{eq:ocp_cost}-\eqref{eq:ocp_initial_variable}, \(N\) is the number of discretization steps, \(\bm{x}_k \in \mathbb{R}^{\Nx}\) and \(\bm{u}_k \in \mathbb{R}^{\Nu}\) denote the state and control input at the \(k\)-th discretization step, respectively.
\eqref{eq:ocp_cost} is the cost function.
\(l_{k}(\cdot, \cdot)\) is the stage cost and \(l_N(\cdot)\) is the terminal cost\revise{; their details are described in \secref{section:cost_and_constraints}}.
\eqref{eq:ocp_state_eq} is the \revise{nonlinear} state \revise{transition that describes the discretized whole-body dynamics, whose concrete form is given in \eqref{eq:state_transition}}.
\eqref{eq:ocp_ineq_con} is the inequality constraint, \revise{which expresses conditions such as the lower and upper bounds of the control input}.
\eqref{eq:ocp_eq_con} is the equality constraint \revise{which expresses conditions such as the contact consistency of the dynamics in \eqref{eq:robot_dynamics}}.
\eqref{eq:ocp_initial_variable} is \revise{the} initial state constraint.
This OCP is solved using numerical optimization algorithms such as sequential quadratic programming (SQP).
In recent years, fast nonlinear optimal control solvers that exploit the sparse structure of the problem have also been developed, enabling real-time MPC applications \cite{mastalli2020crocoddyl, hpipm}.
}{
ここで, \(N\)は離散数, \(\bm{x}_k \in \mathbb{R}^{\Nx}\), \(\bm{u}_k \in \mathbb{R}^{\Nu}\)はそれぞれ\(k\)番目の離散化ステップにおける状態および制御入力を表す.
\eqref{eq:ocp_cost}はコスト関数, \eqref{eq:ocp_state_eq}は状態方程式, \eqref{eq:ocp_ineq_con}は不等式制約, \eqref{eq:ocp_eq_con}は等式制約, \eqref{eq:ocp_initial_variable}は初期状態を表す.
\(l_{k}(\cdot, \cdot)\)はステージコスト, \(l_N(\cdot)\)はターミナルコストを表し, その内訳は\secref{section:cost_and_constraints}で述べる.
この最適制御問題は、非線形計画問題(NLP)に帰着され、逐次二次計画法(SQP)などの数値最適化アルゴリズムを用いて解かれる.
また, 近年, 問題のスパース構造を活用し, リアルタイムMPCとしての適用が可能な高速な非線形最適化ソルバーも開発されている \cite{mastalli2020crocoddyl, hpipm}.
}


\section{Thrust-Rate Input Whole-Body MPC}
\label{sec:thrust_rate_mpc}
\switchlanguage{
In this section, we describe the thrust-rate input whole-body MPC for multi-contact motions.

In conventional multirotor flight control, thrust is directly treated as a control input.
Similarly, in whole-body MPC for articulated robots equipped with thrusters, thrust \(\bm{\lambda}_k \in \mathbb{R}^{\Nr}\) and joint torque \(\bm{\tau}_k \in \mathbb{R}^{\Njoint}\) can be defined as the control input at the \(k\)-th node:
}{
本節では, 多点接触と接触の切替の伴う飛行ヒューマノイドの抗重力歩行のためのThrust-Rate Input Whole-Body MPCについて説明する.
\figref{figure:control_flow}に示す, 提案する最適制御法では, ロータ推力そのものではなく推力の時間微分をMPCの制御入力とし, 推力を状態に含めた定式化を行う.
これにより, 接触切替えに伴う力学的構造の変化に対して, 推力軌道の連続性を状態方程式として保証し, 実機のロータが追従可能な滑らかな推力軌道を生成する.

一般的に, マルチロータ型ロボットの飛行制御では, ロータ推力を直接制御入力とする.
そのため, ロータ搭載多関節ロボットの全身MPCでは, ロータ推力\(\bm{\lambda}_k \in \mathbb{R}^{\Nr}\)と関節トルク\(\bm{\tau}_k \in \mathbb{R}^{\Njoint}\)を同一時刻の入力として扱い,
}
\begin{equation}
    \label{eq:thrust_input_mpc}
    \bm{u}_k^{\mathrm{thrust}} = \begin{bmatrix}
        \bm{\lambda}_k\transpose & \bm{\tau}_k\transpose
    \end{bmatrix}\transpose.
\end{equation}
\switchlanguage{
However, in contact-rich motions such as walking, the contact Jacobian and the dimension of the contact wrench can change discontinuously at contact switching.
As a result, an optimal solution satisfying dynamic equilibrium may require step-like changes in \(\bm{\lambda}_k\).
Real rotors cannot track such thrust spikes instantaneously, and directly applying such commands can cause slipping or detachment during contact switching.
}{
できる.
しかし, 歩行のように, 接触の切り替わる動作を扱う場合には接触ヤコビアンと接触レンチの自由度が不連続に変化するため, 動的平衡を満たす最適解が\(\bm{\lambda}_k\)にステップ状の変化を要求しうる.
実機のロータはこのような推力スパイクを発生できず, そのままでは接触切替え時の滑りや離脱を誘発するという問題がある.
}
\newcommand{\thrustrate}[1]{\dfrac{\mathrm{d}\bm{\lambda}_{#1}}{\mathrm{d}t}}
\newcommand{\thrustrateinline}[1]{\frac{\mathrm{d}\bm{\lambda}_{#1}}{\mathrm{d}t}}
\switchlanguage{
Therefore, in this work, we augment the state with thrust and use the time derivative of the thrust, \(\thrustrateinline{k} \in \mathbb{R}^{\Nr}\), as the input.
We define the augmented state and input as
}{
そこで本研究では, ロータ推力を状態として拡張し, 推力の時間微分\(\thrustrate{k} \in \mathbb{R}^{\Nr}\)を入力とする定式化を行う.
拡張状態と入力を
}
\begin{equation}
    \label{eq:thrust_rate_state_input}
    \bm{x}_k = \begin{bmatrix}
        \bm{q}_k\transpose & \bm{v}_k\transpose & \bm{\lambda}_k\transpose
    \end{bmatrix}\transpose, \quad
    \bm{u}_k = \begin{bmatrix}
        \thrustrate{k}\transpose & \bm{\tau}_k\transpose
    \end{bmatrix}\transpose.
\end{equation}
\switchlanguage{
The generalized force generated by the thrust and joint torque is given by,
}{
と定義する.
このとき, ロータと関節が生成する一般化力は
}
\begin{equation}
    \label{eq:generalized_force_thrust_rate}
    \bm{\tau}^{\mathrm{gen}}_k
    =
    \begin{bmatrix}
        \jacobian{\lambda_1}\transpose
        \begin{bmatrix} \zaxis\\\sigma_1\zaxis\end{bmatrix}
        \cdots
        \jacobian{\lambda_{\Nr}}\transpose
        \begin{bmatrix} \zaxis\\\sigma_{\Nr}\zaxis\end{bmatrix}
    \end{bmatrix}\bm{\lambda}_k
    + \begin{bmatrix}
        \bm{0}_{6}\\
        \bm{\tau}_k
    \end{bmatrix}.
\end{equation}
\switchlanguage{
At each MPC node, the Jacobian of rotor frame \(\jacobian{\lambda_i}\) is calculated from the configuration \(\bm{q}_k\).
With this augmentation, the MPC control input \(\thrustrateinline{k}\) does not directly generate generalized force, but changes the rotor thrust from the next step onward  through integration of the state \(\bm{\lambda}_k\).
}{
で与えられる.
ここでロータフレームのヤコビアン\(\jacobian{\lambda_i}\)はコンフィギュレーション\(\bm{q}_k\)から計算する.
すなわち, MPCにおける制御入力\(\thrustrate{k}\)は直接一般化力を発生せず, 状態\(\bm{\lambda}_k\)の積分を通して次ステップ以降のロータ推力を変化させる.
}

\switchlanguage{
In this work, we use the Euler integration to compute the discrete-time transition of the augmented state as
}{
この拡張状態の離散時間遷移は, Euler積分モデルにより
}
\begin{equation}
    \label{eq:state_transition}
    \bm{x}_{k+1}
    =
    \begin{bmatrix}
        \bm{q}_k \oplus (\bm{v}_k\Delta t + \dot{\bm{v}}_k\Delta t^2)\\
        \bm{v}_k + \dot{\bm{v}}_k\Delta t\\
        \bm{\lambda}_k + \thrustrate{k}\Delta t
    \end{bmatrix}.
\end{equation}
\switchlanguage{
In \eqref{eq:state_transition}, \(\Delta t\) is the discretization time step, and \(\oplus\) denotes the integration operation on \(\mathbb{SE}(3)\).
The generalized acceleration \(\dot{\bm{v}}_k\) is calculated by forward dynamics\revise{.}
}{
計算する.
\eqref{eq:state_transition}において,\(\Delta t\)は離散化時間, \(\oplus\)は\(\mathbb{SE}(3)\)上の積分演算である.
また, 一般化加速度\(\dot{\bm{v}}_k\)は接触拘束を満たす順動力学
}
\switchlanguage{
From \eqref{eq:state_transition}, thrust evolves by integrating thrust-rate and therefore changes continuously across MPC nodes.
With a sufficiently small  \(\Delta t\), this yields a smooth thrust trajectory that real rotors can track even when the dynamic structure changes due to contact switching.
Furthermore, since we do not add constraints that directly couple thrust variables at consecutive nodes \(\qty(\bm{\lambda}_{k+1}-\bm{\lambda}_{k})\), the sparse structure of the OCP is preserved.
Therefore, not only general-purpose NLP solvers but also fast SQP solvers that exploit the sparse structure can be applied, enabling real-time MPC applications.
}{
により計算する.
式\equref{eq:state_transition}より, 推力は各サンプリング区間で線形に変化するため, 離散化時間\(\Delta t\)を十分小さくすることで推力の連続性が状態方程式として保証される.
さらに, 連続する二つの入力\(\bm{\lambda}_{k+1}-\bm{\lambda}_{k}\)を直接結ぶrate制約をOCPへ追加しないため, 問題のスパース性は維持される.
そのため, 汎用NLPソルバーのみでなく, 現代の最適制御ソルバーが利用するスパース構造を活用した高速なSQPソルバーが利用可能で, リアルタイムでのMPC適用が可能になる.
}

\section{Anti-Gravity Walking MPC Formulation}
\label{sec:anti_gravity_walking_mpc}
\switchlanguage{
In this section,
we formulate the OCP for anti-gravity walking.
\revise{Among the inequality constraints \(\bm{c}\) in \eqref{eq:ocp_ineq_con}, bounds of the control input \(\bm{u}_k\) are imposed as box constraints handled by the solver, whereas the CWC condition is treated as a soft penalty in the cost function.
This is because imposing the CWC as a hard inequality constraint requires a constrained solver, which increases computation time per iteration.
We prioritize real-time performance, and make the soft penalty sufficient by giving appropriate references for the swing foot and CoM trajectories.
}}{
この節では, \secref{sec:thrust_rate_mpc}で提案した推力差分入力型全身MPCをベースに, 抗重力歩行のための最適制御問題を定式化する.
\eqref{eq:ocp_ineq_con}の不等式制約\(\bm{c}\)のうち, 制御入力\(\bm{u}_k\)の上下限はソルバーが扱えるハードな箱制約として課し, 接触レンチ錐条件はコスト関数中のソフトなペナルティとして扱う.
これは, 接触レンチ錐をハードな不等式制約として課すには制約付きソルバーが必要となり, 1反復あたりの計算量と収束までの反復数が増加するためである.
我々は実時間性を優先し, 遊脚軌道とCoM軌道に妥当な参照を与えることでソフトペナルティでも扱えるようにした.}

\subsection{Cost and Constraints for Anti-Gravity Walking}
\label{section:cost_and_constraints}
\switchlanguage{
\revise{\revise{T}he stage cost \(l_k\) in \eqref{eq:ocp_cost} is expressed as follows,}
}{
\revise{抗重力歩行の全身MPCでは, \eqref{eq:ocp_cost}のステージコスト\(l_k\)を以下の項から構成する.}}
\revise{\begin{equation}
    \label{eq:stage_cost}
    l_k(\bm{x}_k, \bm{u}_k)
    = l_{\mathrm{foot},k}
    + l_{\mathrm{CoM},k}
    + l_{\mathrm{CWC},k}
    + l_{\mathrm{thrust},k}
    + l_{\mathrm{reg},k}.
\end{equation}}%
\switchlanguage{\revise{In \eqref{eq:stage_cost}, \(l_{\mathrm{foot},k}\) and  \(l_{\mathrm{CoM},k}\) are the swing foot and CoM tracking costs that generate the walking motion along the contact surface.
\(l_{\mathrm{CWC},k}\) is the penalty for the contact wrench cone condition.
\(l_{\mathrm{thrust},k}\) regularizes the thrust and penalizes its saturation, and \(l_{\mathrm{reg},k}\) is the state and input regularization term.
The terminal cost \(l_N(\bm{x}_N)\) consists of the same terms as the stage cost except for the input regularization.
In the following, \(\norm{\bm{a}}^2_{\bm{Q}} = \bm{a}\transpose\bm{Q}\bm{a}\) denotes the weighted squared norm.
We also define \(\bm{\delta}\) that denotes the deviation of \(\bm{a}\) from its bounds \(\underline{\bm{b}}, \bar{\bm{b}}\),}
}{
\eqref{eq:stage_cost}において, \(l_{\mathrm{CoM},k}\), \(l_{\mathrm{foot},k}\)は接触面に沿った歩行動作を生成する重心・遊脚追従コスト, \(l_{\mathrm{CWC},k}\)は接触レンチ錐(CWC)制約のペナルティである.
\(l_{\mathrm{thrust},k}\)は推力の正則化とsaturationのペナルティ, \(l_{\mathrm{reg},k}\)は状態・入力の正則化項である.
ターミナルコスト\(l_{N}(\bm{x}_{N})\)では後述する入力正則化以外の項を含めた.
}
\revise{\begin{equation}
    \label{eq:deviation}
    \bm{\delta}\qty(\bm{a}; \underline{\bm{b}}, \bar{\bm{b}})
    = \max\qty(\bm{0}, \bm{a}-\bar{\bm{b}}) + \min\qty(\bm{0}, \bm{a}-\underline{\bm{b}}).
\end{equation}}%
\switchlanguage{\revise{Here, \(\max\) and \(\min\) are applied elementwise, so \(\bm{\delta}\) is zero for the components within the bounds and gives the signed deviation otherwise.
In the following, each term in \eqref{eq:stage_cost} is described.
}}{\revise{
ここで\(\max, \min\)は要素ごとに作用し, \(\bm{\delta}\)は上下限内の成分に対してはゼロ, それ以外では符号付きの逸脱量を与える.
\eqref{eq:stage_cost}の各項は, 対応する残差の重み付き二乗ノルムとして与える.}}

\subsubsection{\revise{Swing Foot} and \revise{CoM} Tracking}
\label{section:com_and_swing_foot_tracking}
\switchlanguage{
Swing foot reference trajectories are generated by linearly interpolating in walking direction based on the target velocity given by the motion planner.
The foot-lift motion is generated by a sinusoidal curve in the direction perpendicular to the contact surface.
\revise{Let \(\bm{r}_{\mathrm{foot},k} \in \mathbb{R}^{6}\) be the pose error of the swing foot from this reference, then \(l_{\mathrm{foot},k}\) is calculated as \(\norm{\bm{r}_{\mathrm{foot},k}}^2_{\bm{Q}_{\mathrm{foot}}}\) with a weight matrix \(\bm{Q}_{\mathrm{foot}} \in \mathbb{R}^{6\times6}\).}

The CoM reference trajectories are set to keep the CoM height \revise{from the contact surface}.
During the swing phase, it is set above the support foot, and during the double-support phase, it transitions to a position above the next support foot by linear interpolation.
\revise{Similarly, with the CoM tracking error \(\bm{r}_{\mathrm{CoM},k} \in \mathbb{R}^{3}\), \(l_{\mathrm{CoM},k}\) is calculated as \(\norm{\bm{r}_{\mathrm{CoM},k}}^2_{\bm{Q}_{\mathrm{CoM}}}\) with a weight matrix \(\bm{Q}_{\mathrm{CoM}} \in \mathbb{R}^{3\times3}\).}
}{
足平の参照軌道は, 動作計画器から与えられた目標歩行速度に基づき, 進行方向に線形補間することで生成する.
足上げ方向は接触面に垂直な方向に\(\sin\)カーブで生成する.
これを\(\bm{Q}_{\mathrm{foot}}\)で重み付けし, 計算した\(l_{\mathrm{foot}, k}\)コストを追加する.
CoMの参照軌道は重心高さを維持しつつ, 遊脚期には支持脚の上方に設定し, 両足支持期において次の支持脚の上方に線形補間で遷移させる.
同様に\(\bm{Q}_{\mathrm{CoM}}\)で重みつけする.
}

\subsubsection{Contact Wrench Cone \revise{Penalty}}
\label{section:contact_wrench_cone}
\switchlanguage{
During walking, it is necessary to prevent feet from slipping and rolling around its edge.
This condition can be expressed by CWC \revise{constraints} defined in \cite{caron2015stability}.
Let \(\bm{w}_{c_{j}, k} = [f_{{j_x},k}, f_{{j_y},k}, f_{{j_z},k}, \tau_{{j_x},k}, \tau_{{j_y},k}, \tau_{{j_z},k}]\) be the contact wrench at the \(j\)-th foot, \revise{then  the CWC constraint is written as \(\underline{\bm{b}}_{j, k} \le \bm{A}_{j, k} \bm{w}_{c_{j}, k} \le \bar{\bm{b}}_{j, k}\), where \(\bm{A}_{j, k} \in \mathbb{R}^{m\times 6}\) is a matrix determined by the friction coefficient \(\mu\) and the foot size.
Here, \(\underline{\bm{b}}_{j, k}, \bar{\bm{b}}_{j, k} \in \mathbb{R}^{m}\) are the lower and upper bounds of these constraints.}
}{
歩行動作において, 足平が滑ることや境界にそって転がることを防ぐ必要がある.
この条件を, 足平における接触レンチが\cite{caron2015stability}で定義される接触レンチ錐の内部にあることを要求することで表す.
\(j\)番目の足平における接触レンチを\(\bm{w}_{c_{j}, k} = [f_{j_{x}, k}, f_{j_{y}, k}, f_{j_{z}, k}, m_{j_{x}, k}, m_{j_{y}, k}, m_{j_{z}, k}]\transpose\)とすると, この条件は以下のように表される.
}
\switchlanguage{

In anti-gravity walking, natural support forces from gravity are not available, and the foot must be actively pressed against the contact surface by thrust.
However, since a thrust-minimization term is included \revise{in} the cost function, the resulting contact \revise{force} can become excessively small, \revise{and} the foot may detach due to disturbances even if the CWC constraint is satisfied.
Therefore, we introduce a lower bound \(f_{j_{z}, k}^{\min} \in \mathbb{R}_{+}\) on the foot-normal component of the contact force, \revise{whose setting} is described in \secref{sec:min_normal_force_transfer}.
\revise{Using \(\bm{\delta}\) in \eqref{eq:deviation}, the penalty for the \(\Nc\) contact points is as follows, with a weight matrix \(\bm{Q}_{\mathrm{CWC}} \in \mathbb{R}^{m\times m}\):}
}{
ここで\(\bm{A}_{j, k} \in \mathbb{R}^{m\times 6}\)は摩擦係数\(\mu\)と接触面の幾何形状により定まる\(j\)番目の接触点における接触レンチ錘を表す行列であり, \(\underline{\bm{b}}_{j, k}, \bar{\bm{b}}_{j, k} \in \mathbb{R}^m\)はそれぞれ接触レンチ錐の下限と上限を表すベクトルである.
抗重力歩行では, 重力による自然な床反力が得られず, ロータ推力によって足裏を接触面へ能動的に押し付ける必要がある.
接触レンチ錘の制約を満たす解が得られたとしても, 十分な接触力が発生していない場合, 外乱などによって足平が遊離する恐れがある.
さらに推力を最小化するコストを入れて最適化すると過剰に小さな接触レンチになることが予期される.
よって本研究では,  接触力の足平法線方向の成分について, 下限値を\(f_{j_{z}, k}^{\min} \in \mathbb{R}_{+}\)として導入する.
この最小接触力\(f_{j_{z}, k}^{\min}\)の設定方法については\secref{sec:min_normal_force_transfer}で説明する.
各接触点に対して, 接触レンチ錘から逸脱した成分に対して\(\bm{Q}_{\mathrm{CWC}}\)で重み付けしたペナルティ項を\(l_{\mathrm{CWC},k}\)として加える.
}
\revise{\begin{equation}
  l_{\mathrm{CWC},k} = \sum_{j=1}^{\Nc} \norm{\bm{\delta}\qty(\bm{A}_{j, k} \bm{w}_{c_{j}, k}; \underline{\bm{b}}_{j, k}, \bar{\bm{b}}_{j, k})}^2_{\bm{Q}_{\mathrm{CWC}}}.
\end{equation}}%

\subsubsection{Thrust Regularization and \revise{Saturation Penalty}}
\label{section:thrust_regularization}
\switchlanguage{
To reduce energy consumption, it is desired to minimize the generated thrust.
In addition, thrust has lower and upper limits due to hardware limitations. 
\revise{Therefore,} the cost \revise{related to thrust} at the \(k\)-th node is expressed as follows,
}{
消費エネルギを低減すべく, 発揮推力は必要最小限に抑えることが望ましい.
また, ハードウェア的にロータ推力には最小・最大値が存在するため, これらの制約を考慮する必要がある.
}
\revise{
\begin{equation}
  \label{eq:thrust_regularization}
  l_{\mathrm{thrust},k} = \norm{\bm{\lambda}_k}^2_{\bm{Q}_{\mathrm{thrust}}} + \norm{\bm{\delta}\qty(\bm{\lambda}_k; \underline{\bm{\lambda}}, \bar{\bm{\lambda}})}^2_{\bm{Q}_{\mathrm{limit}}}.
\end{equation}}%
\switchlanguage{In \eqref{eq:thrust_regularization}, \(\bar{\bm{\lambda}}, \underline{\bm{\lambda}} \in \mathbb{R}^{\Nr}\) denote the upper and lower limits of rotor thrust, respectively, and \(\bm{Q}_{\mathrm{thrust}}, \bm{Q}_{\mathrm{limit}} \in \mathbb{R}^{\Nr\times\Nr}\) are the weight matrices for thrust regularization and saturation penalty, respectively.
}{
ここで, \(\bar{\bm{\lambda}}, \underline{\bm{\lambda}} \in \mathbb{R}^{\Nr}\)はそれぞれロータ推力の上限と下限を表すベクトルであり, \(\bm{Q}_{\mathrm{thrust}}, \bm{Q}_{\mathrm{limit}} \in \mathbb{R}^{\Nr\times\Nr}\)はそれぞれ推力正則化と制約ペナルティの重み行列である.
}

\subsubsection{State and Input Regularization}
\label{section:state_and_input_regularization}
\switchlanguage{
To enhance the physical validity of the optimal solution, we also add regularization terms for the state and input.
\revise{Let \(\bm{x}_{\mathrm{ref}}\) be the reference state, which is a standing pose with the knee slightly bent and zero velocity, and let \(\bm{r}_{x,k} \in \mathbb{R}^{2\Nv}\) be the error of \(\bm{x}_k\) from \(\bm{x}_{\mathrm{ref}}\) except for the thrust.
Then, \(l_{\mathrm{reg},k}\) is \(\norm{\bm{r}_{x,k}}^2_{\bm{Q}_{x}} + \norm{\bm{u}_k}^2_{\bm{Q}_{u}}\) with weight matrices \(\bm{Q}_{x} \in \mathbb{R}^{2\Nv\times2\Nv}\) and \(\bm{Q}_{u} \in \mathbb{R}^{\Nu\times\Nu}\), which suppresses the deviation from the reference state, thrust change,} and the magnitude of the joint torque.
}{
最適解の物理的妥当性を高めるため, 状態と入力についても正則化項を加える.
状態正則化は, 膝を少し曲げた立位時のコンフィギュレーションと, ゼロ速度を参照にし, 現在状態とのエラーを\(\bm{Q}_x \in \mathbb{R}^{2\Nv\times2\Nv}\)で重みつけし, \(l_{\mathrm{reg},k}\)に加える.
入力正則化は, ロータ推力の変化量と関節トルクの大きさを抑えるため, \(\bm{u}_{k}\)を\(\bm{Q}_u \in \mathbb{R}^{\Nu\times\Nu}\)で重みつけし, \(l_{\mathrm{reg},k}\)に加える.
}

\newcommand{\leftletter}{\text{L}}
\newcommand{\rightletter}{\text{R}}
\subsection{Load Transfer Strategy}
\label{sec:min_normal_force_transfer}
\begin{figure}[t]
  \centering
  \includegraphics[width=1.0\columnwidth]{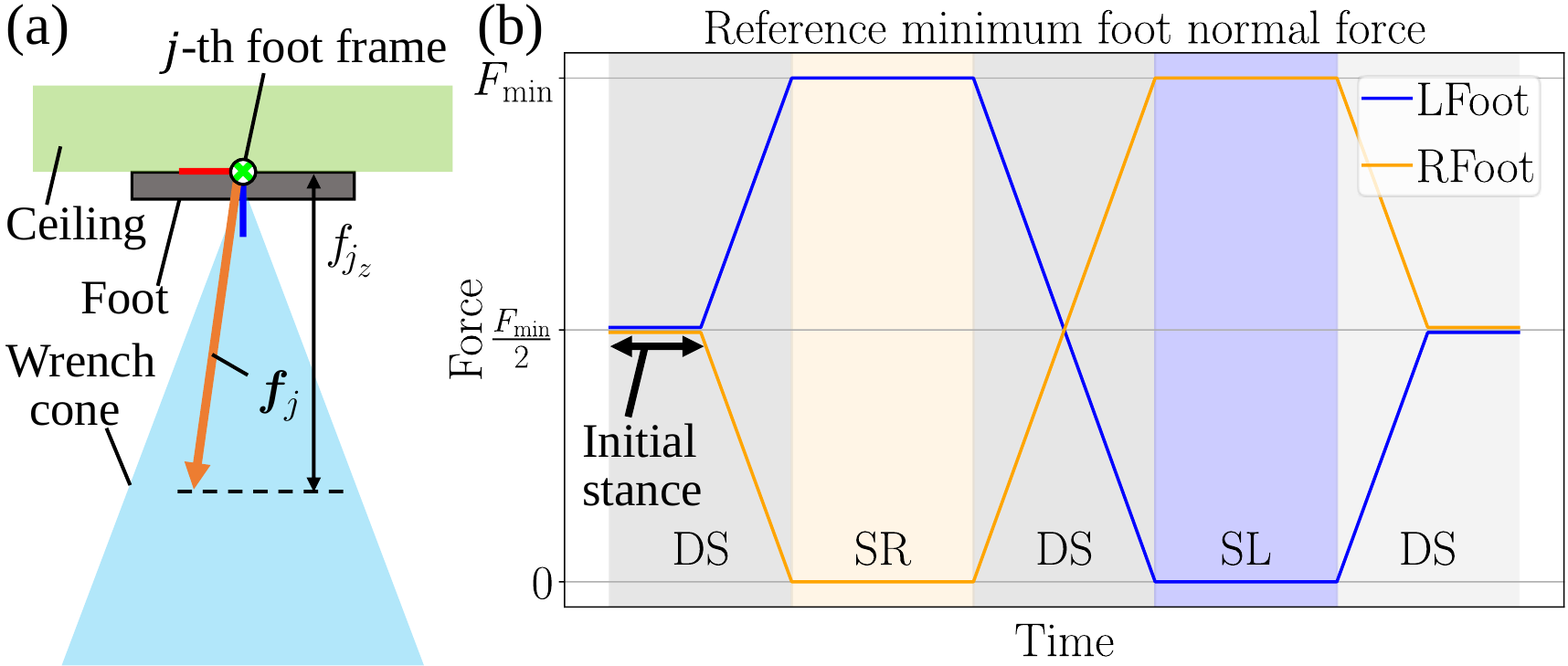}
  \caption{(a) Contact frame and normal force \(f_{j_{z}}\) definition. (b) Reference minimum normal force trajectory. \revise{\(F_{\min}\) is the minimum normal force during double support phase.
      DS, SR, and SL denote the double support, right swing, and left swing phases, respectively.}}
    \vspace{-2mm}
  \label{figure:minimum_normal_force_transfer}
\end{figure}

\switchlanguage{
\revise{D}uring walking, the foot contact state switches between double-support and single-support, and simply assigning a constant value to the support foot causes a discontinuous change in \(f_{j_{z}, k}^{\min}\) at contact switching, \revise{which can} make the optimal solution require a sudden change in the target thrust.
Therefore, as shown in \figref{figure:minimum_normal_force_transfer}, we propose a method to smoothly switch these bounds during the double-support phase.
In this work, we assume that contact occurs at feet and that the robot takes either double-support or single-support phases.
The minimum contact force \(F_{\min}\) is assigned only to the support foot during the single-support phase.
Let the right foot swing phase be \(\phi=\mathrm{SR}\) and the left foot swing phase be \(\phi=\mathrm{SL}\).
Then, the minimum foot-normal component of contact force for \revise{each} foot can be expressed as follows,
}{
抗重力環境において十分な接触力を発生させるべく, 接触レンチ錐の下限に最小押し付け力\(f_{j_{z}, k}^{\min}\)を設定することを\secref{section:contact_wrench_cone}で述べた.
しかし, 歩行動作では足の接触状態が両足支持から片足支持へ, あるいはその逆へと切り替わるため, 支持脚に対して一定値を与えるだけでは, 接触切替えの前後で法線力参照が不連続に変化してしまう.
その結果, 目標発揮推力の急激な変化を要求する最適解が生成される恐れがある.
よって, \figref{figure:minimum_normal_force_transfer}に示すように, 両足支持期を利用してこの最小接触力を滑らかに切り替える方法を提案する.

本研究では飛行ヒューマノイドにおいて足平において接触が発生し, 両足支持または片足支持の状態を取ると仮定する.
最小押し付け力\(F_{\min}\)は, 片脚支持期では支持足のみに与える.
右足遊脚期を\(\phi=\mathrm{SR}\), 左足遊脚期を\(\phi=\mathrm{SL}\)とすると,
}
\begin{equation}
    \qty(f_{L_{z}}^{\min}, f_{R_{z}}^{\min})
     = \begin{cases}
        (F_{\min}, 0) & \phi=\mathrm{SR},\\
        (0, F_{\min}) & \phi=\mathrm{SL}.
    \end{cases}
\end{equation}
\switchlanguage{
Here\revise{,} \ \(L=1\) and \(R=2\) denote the left and right foot, respectively, and \(f_{L_{z}}^{\min}, f_{R_{z}}^{\min}\) represent the minimum foot-normal component of contact forces for the left and right foot, respectively.
In the double-support phase, for the number of nodes in the double-support phase \(N_{\mathrm{DS}}\), we define a progress ratio \(\rho \in [0,1]\) for each node and linearly transfer these bounds between the left and right feet.
This can be formulated as follows,
}{
ただし, \(L=1\), \(R=2\)とし, \(f_{L_{z}}^{\min}, f_{R_{z}}^{\min}\)はそれぞれ左足と右足の最小法線力を表す.
両脚支持期では, 支持脚の切替えに伴う接触力の不連続を避ける必要がある.
そのため, 両足支持期のステップ数\(N_{\mathrm{DS}}\)に対して, 各ノードの進行率\(\rho \in [0,1]\)を定義して左右の最小接触力を線形に遷移させる.
これを定式化すると以下のようになる.
}
\begin{equation}
    \label{eq:min_normal_force_transfer}
    \begin{bmatrix}
        f_{L_{z}}^{\min}\\
        f_{R_{z}}^{\min}
    \end{bmatrix}
    = F_{\min}
    \qty((1-\rho)
    \bm{\eta}^{\mathrm{last}}
    + \rho
    \bm{\eta}^{\mathrm{next}}).
\end{equation}
\switchlanguage{
In \eqref{eq:min_normal_force_transfer}, \(\bm{\eta}^{\mathrm{last}}, \bm{\eta}^{\mathrm{next}} \in \mathbb{R}^2\) are one-hot vectors representing the support foot just before entering the double-support phase and the foot that will become the support foot in the next swing phase, respectively.
With this load transfer strategy, the reference of foot-normal component of contact forces changes continuously at contact switching, enabling walking that is consistent with continuous thrust generation by the thrust-rate input MPC.
Note that this minimum contact force \(F_{\min}\) must be appropriately chosen according to the physical quantities and friction conditions of the robot to which the controller is applied; in this work, it was determined experimentally.
}{
ここで\(\bm{\eta}^{\mathrm{last}}, \bm{\eta}^{\mathrm{next}} \in \mathbb{R}^2\)はそれぞれ両脚支持期へ入る直前の支持足, は次の遊脚期で支持足となる足を表すone-hotベクトルである.
この荷重移動により, 接触切替えの前後で法線力参照が滑らかに変化し, 推力差分入力による連続推力生成と整合した歩行が可能になる.
なお, この最小接触力\(F_{\min}\)は制御を適用するロボットの慣性や摩擦条件に応じて適切な値を選ぶ必要があり, 今回は実験的に求めた.
}

\subsection{Real-Time MPC Implementation}
\label{section:rti}
\switchlanguage{
To implement the proposed thrust-rate input whole-body MPC, we extended Crocoddyl \cite{mastalli2020crocoddyl} based on \revise{\eqref{eq:robot_dynamics}} and \eqref{eq:thrust_rate_state_input}--\eqref{eq:state_transition} so that it can handle articulated robots with arbitrary thruster distributions.
The optimization is warm-started using \revise{the latest solution} whose initial state is updated with the current state.
To solve the OCP, we use a BoxFDDP solver \revise{\cite{boxfddp}} that can handle the upper and lower bounds on thrust-rate and joint torques, and update the solution with one iteration\revise{, and send the latest command.}
We perform parallel computation with Crocoddyl's multithreading feature.

\revise{
The horizon is advanced by one node based on the elapsed wall-clock time: the node shift is triggered only after the elapsed time exceeds the MPC discretization step \(\Delta t\).
Within each node interval, the solver is executed repeatedly, and the latest solution is applied as the control command.
This decouples the node shift, which must be consistent with the model's time discretization, from the per-cycle computation time.
Consequently, even when a single solve occasionally exceeds the control cycle, as long as it stays below \(\Delta t\), the temporal alignment between the horizon and real time is preserved and no node is skipped.
}

In addition, when the first node of the horizon enters the double-support phase, we project the actual foot position onto the contact surface and regenerate the foot reference and CoM references.
This suppresses the accumulation of small contact slip and model errors over the horizon.
}{
動力学計算にはPinocchio\cite{carpentier2019pinocchio}を用い, 最適制御問題はCrocoddyl\cite{mastalli2020crocoddyl}上に実装した.
なお, 本研究で提案する推力差分入力型全身MPCのため, \eqref{eq:thrust_rate_state_input}--\eqref{eq:thrust_rate_forward_dynamics}に基づきCrocoddylを拡張し, 任意のロータ分散型多関節ロボットを扱えるようにした.
最適化は前制御周期の解を一ノード分シフトし, 現在の状態で\(\bm{q}_0, \bm{v}_0\)を更新した\(\statetraj, \inputtraj\)を初期解としてwarm-startする.
求解には推力差分入力と関節トルクの上下限制約を考慮できるBoxFDDPソルバーを用いて, 1イタレーションの計算で解を更新する.
CrocoddylのMulti-threading機能を利用し, 4threadで並列計算を行った.
制御出力としては, 関節には次ノードの関節位置・速度および最適トルクを送り, ロータには次ノードの推力状態\(\bm{\lambda}_1\)を送る.

ホライズンのノードシフトは, 経過した実時間がMPC離散化ステップ\(\Delta t\)を超えたときに行う.
各ノード区間では, 前解をwarm-startとして繰り返し求解し, 最新解を制御指令として適用する.
これにより, モデルの時間離散化と整合させるべきノードシフトを, 周期ごとの計算時間から分離できる.
その結果, 単一の求解がの制御周期を稀に超えても, \(\Delta t\)未満であればホライズンと実時間の整合は保たれ, ノードのスキップも生じない.

また, ホライズン先頭が両脚支持へ入ったタイミングでは, 実際の足先位置を接触面へ射影し, 以降の足先参照とCoM参照を再生成する.
これにより, 接触中の微小な滑りやモデル誤差がホライズン全体に蓄積することを抑える.
}

\renewcommand{\arraystretch}{0.7}
\begin{table}[t]
	\centering
	\caption{Specifications of the robot model}
	\label{table:robot_spec}
	\begin{tabular}{|c|c|c|}
		\hline
		Parameter & Value & Unit\\
		\hline
		Robot height & 0.4 & m\\
		\revise{M}ass in \revise{Simulation / Hardware} & 1.6 / 2.0 & kg\\
		Max joint torque & 1.8 & Nm\\
		Max thrust & 20.0 & N\\
		Friction coefficient & 0.7 & \\
		Step height (used in \secref{section:com_and_swing_foot_tracking})& 0.03 & m\\
		Foot size (used in \secref{section:contact_wrench_cone}) & 0.1 \(\times\) 0.\revise{0}6 & m \(\times\) m\\
		Time step & 0.025 & s\\
        \revise{MPC Horizon} & \revise{1.0} & \revise{s}\\
		Swing foot duration & 0.775 & s\\
		\hline
	\end{tabular}
\end{table}

\renewcommand{\arraystretch}{0.7}
\begin{table}[t]
 \centering
 \caption{Weights for optimization.}
 \label{table:ddp_weight}
 \begin{tabular}[t]{|c|c|c|}
     \hline
	 Param & Value & ref.\\\hline
	 \(\bm{Q}_{\text{foot}}\) & \(\text{diag}\qty(10^6, 10^6, 10^6, 10^5, 10^5, 10^5)\) & \secref{section:com_and_swing_foot_tracking}\\
	 \(\bm{Q}_{\text{CoM}}\) & \(\text{diag}\qty(10^3, 10^4, 10^3)\) & \secref{section:com_and_swing_foot_tracking}\\
	 \(\bm{Q}_{\text{CWC}}\) & \(10^2 \bm{I}_{m} \) & \secref{section:contact_wrench_cone}\\
         \(\bm{Q}_{x}\)&  \(\text{diag}\qty(\bm{w}_{\text{root}, q}^{\transpose}, \bm{w}_{\text{joint}, q}^{\transpose}, \bm{w}_{\text{root}, v}^{\transpose}, \bm{w}_{\text{joint}, v}^{\transpose})\) & \secref{section:state_and_input_regularization}\\
	 \(\bm{w}_{\text{root}, q}\) & \(\qty[0, 0, 0, 10^3, 10^4, 10^2]\) & \\
	 \(\bm{w}_{\text{joint}, q}\) & \(10^{-2}\times \bm{1}_{N_{\text{joint}}}\) & \\
	 \(\bm{w}_{\text{root}, v}\) & \(10\times \bm{1}_{6}\) & \\
	 \(\bm{w}_{\text{joint}, v}\) & \(10^{-1} \times \bm{1}_{N_{\text{joint}}}\) & \\
	 \(\bm{Q}_{u}\) &  \(10 \times \bm{I}_{N_{\text{r}} + N_{\text{joint}}}\) & \secref{section:state_and_input_regularization}\\
     \(\bm{Q}_{\text{thrust}}\) &  \(10^{-2}\times \bm{I}_{N_{\text{r}}}\) & \eqref{eq:thrust_regularization}\\
     \(\bm{Q}_{\text{limit}}\) &  \(10\times \bm{I}_{N_{\text{r}}}\) & \eqref{eq:thrust_regularization}\\
     \hline
 \end{tabular}
\end{table}

\section{Evaluation}
\label{sec:evaluation}
\switchlanguage{
In this section, we verify the effectiveness of the proposed method\revise{s} through trajectory optimization \revise{(TO)}, dynamics simulation\revise{, and hardware experiment}.
}{
本節では, 軌道最適化と動力学シミュレータを用いた評価により, 提案手法の有効性の検証と抗重力歩行動作の実証を行う.
}

\subsection{Trajectory \revise{O}ptimization}
\label{section:trajectory_optimization}
\subsubsection{Problem setup}
\switchlanguage{
First, to verify the effectiveness of the thrust-rate input \revise{OCP} formulation, we performed offline \revise{TO} for ceiling walking.
We compared the results with those obtained by solving the same \revise{TO} problem using thrust input \revise{formulation}, which treats thrust and joint torques as inputs as introduced in \eqref{eq:thrust_input_mpc}.
Robot specifications and common parameters for both optimizations are shown in \revise{\figref{figure:dynamics},} \tabref{table:robot_spec}\revise{,} and \tabref{table:ddp_weight}.
The double-support duration was set to \SI{0.2}{s}.
The walking motion, consisting of four steps, is formulated as a single OCP.
The initial guess for the state was set to a nominal joint configuration with both feet in contact with the ceiling and the knees bent.
Moreover, thrust vect\revise{o}ring angles of each thruster was set to make thrust direction upward.
The initial guess for the thrust-rate was set to zero, and the initial guesses for the thrust and joint torques \revise{are computed by QP to satisfy static equilibrium with contact forces and gravity}.
Here, to evaluate the formulation of the OCP with thrust-rate as input, the minimum contact force \(F_{\text{min}}\) was set to \SI{0}{N}.
The computation was performed on a standard laptop with an Intel Core i7-10850H@\SI{2.7}{GHz}, \revise{and parallelized with four threads}.
}{
まず, 推力差分入力型全身MPCの有効性を検証すべく, 天井での抗重力歩行動作の軌道最適化を行う.
同じ動作の軌道最適化問題を, \eqref{eq:thrust_input_mpc}で紹介したような, 推力と関節トルクを入力とするThrust input MPCで解く場合と比較する.
両足支持時間は\SI{0.2}{s} (8 time steps)に設定した.
これは4歩の歩行動作からなる計6.65秒の動作であり, これを1つの最適制御問題として定式化し, 最大100回のイテレーションで最適解を求める.
状態に関する初期解は両足平を天井に接触させ, 膝を曲げた標準関節角度にする.
また, 推力差分の初期解はゼロベクトル, 推力と関節トルクの初期解は重力, 接触力と釣り合うように, 二次計画法によって求める.
なおここではThrust-rateを入力にする最適制御問題の定式化を評価するため, 目標の法線方向の最小接触力は\SI{0}{N}に設定した.
計算は一般的なラップトップ with Intel Core i7@\SI{2.7}{GHz}上で行った.
}



\subsubsection{Results}
\begin{figure}[t]
    \centering
    \includegraphics[width=1.0\columnwidth]{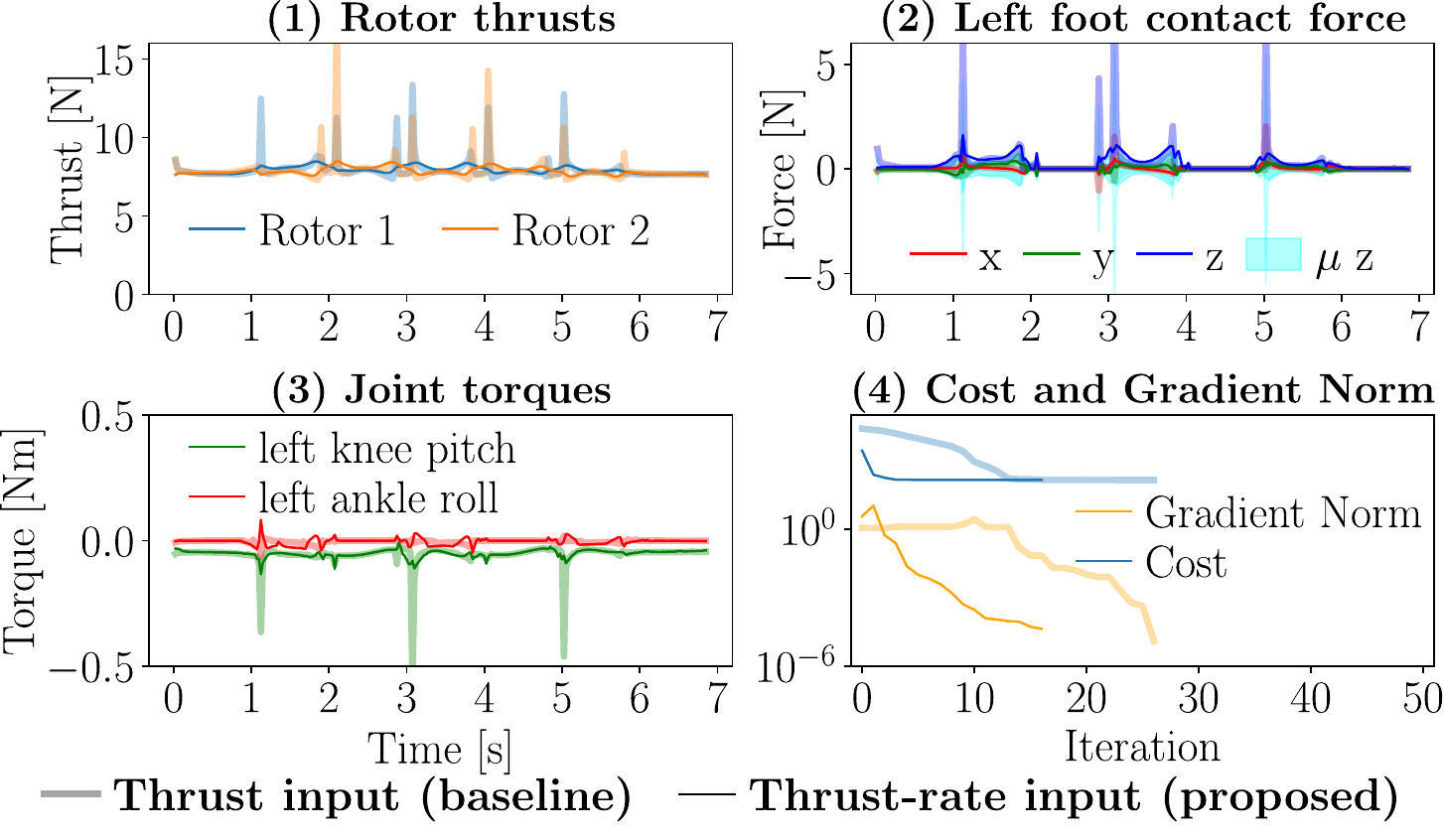}
    \caption{\revise{Comparison of TO results.}}
    \vspace{-3mm}
    \label{figure:trajectory_optimization_thrust_rate}
\end{figure}

\switchlanguage{
The optimal trajectories of the anti-gravity walking motion\revise{s} \revise{are} shown in \revise{the supplemental video}.
The \revise{comparison of} thrust-rate and thrust input formulation\revise{s} \revise{is} shown in \figref{figure:trajectory_optimization_thrust_rate}.
These plots show the rotor thrust, the contact force \revise{of} the left foot, the joint torques of the left knee pitch and ankle roll, and the total cost and its gradient norm \revise{at each iteration}\revise{,} for the obtained optimal trajectories.
In conventional thrust input formulation, step-like changes of about \SI{3}{N} appeared in the thrust during contact switching, whereas in thrust-rate input formulation, the thrust changed smoothly.
This shows that thrust-rate input formulation could suppress thrust spikes at contact switching.
Similar behavior can also be observed in the foot contact force and the leg joint torques.
For the normal component of the foot contact force, thrust input formulation produced a spike of about \SI{5}{N}, and the solution required a spike of about \SI{0.5}{Nm} at knee joint torque.
Moreover, while the thrust input formulation required 26 iterations for optimization to converge, the thrust-rate input formulation converged in 16 iterations.
This is considered to be because thrust-rate input formulation generated a more physically plausible trajectory and improved the numerical stability of the optimization.
These results demonstrate the effectiveness of the proposed formulation of the OCP with thrust-rate as input.
}{
Thrust-rate Input MPCを用いて生成された抗重力歩行動作の軌道をを\figref{figure:trajectory_optimization_snapshots}に示す.
また, Thrust-rate Input MPC, 通常のMPCを用いたときの結果をそれぞれ\figref{figure:trajectory_optimization_thrust_rate}, \figref{figure:trajectory_optimization_thrust}に示す.
これらのプロットは得られた最適軌道におけるロータ推力, 左足の接触力, 左足の膝pitchとangle rollの関節トルク, および最適化におけるコストとその勾配のノルムを示している.
Thrust-Input MPCでは接触の遷移に伴って推力に\SI{3}{N}程度のステップ状の変化が生じているのに対し, Thrust-rate Input MPCでは推力が連続的に変化していることがわかる.
これにより, Thrust-rate Input MPCが接触切替え時の推力スパイクを抑制できることが示された.
同様の現象は足平での接触力や関節トルクにも見られた.
足平接触力の法線方向成分については, 通常のThrust-Input MPCでは接触の切り替わりの瞬間に\SI{5}{N}程度のスパイクが, 膝関節には\SI{0.5}{Nm}程度のそれを要求する解が得られた.
また, Thrust-input MPCでは最適化の収束に26回のイタレーションを要しているのに対し, Thrust-rate Input MPCでは16回のイタレーションで収束した.
これは, Thrust-rate Input MPCの方がより物理的に妥当な軌道を生成するため, 最適化計算の数値的安定性が向上したためであると考えられ, 提案手法の有効性を示した.
}

\subsection{Anti-Gravity walking simulation via MPC}
\label{section:mpc_simulation}
\subsubsection{Problem Setup}
\switchlanguage{
Based on the proposed method, we constructed a whole-body MPC framework and verif\revise{ied} through physics simulation.
\revise{As control outputs, we send the joint positions, velocities, and torques at the next node to the joints, and the thrust state at the next node to the thrusters.}
We performed simulations in a MuJoCo \cite{todorov2012mujoco} environment with a ceiling \revise{using the same computational environment in \secref{section:trajectory_optimization}}.
To allow gradual load transfer, we set the double-support duration to \SI{0.75}{s}.
The control frequency was set to \SI{100}{Hz}.
Noise was added to torso link state obtained from MuJoCo, and the torso link state was estimated by fusing this noisy state and IMU data using an extended Kalman filter.
Considering the robot mass and the maximum available thrust, we set the minimum contact force \(F_{\min}\) as \SI{5}{N}.
}{
提案手法に基づき, 全身MPCを構築し, シミュレーションにより抗重力歩行動作が可能なことを検証する.
天井を用意したMuJoCo\cite{todorov2012mujoco}環境でシミュレーションを行う.
荷重移動をゆっくりにするため, 両足支持期を\SI{0.75}{s} (30 time steps)に設定した.
制御周期は\SI{100}{Hz}に設定した.
MuJoCoから得たルートリンクの状態にノイズを加えたものと, IMUのデータを拡張カルマンフィルタを用いて統合してルートリンクの状態を推定した.
ロボットの質量と発揮可能な最大推力を考慮して, 最小法線力\(F_{\min}\)は\SI{5}{N}に設定した.
初期姿勢の状態で天井に両足平を接触させた状態で一度のみ全軌道を計算し, それ以降は現在の状態を初期解にして1イテレーションのみ解く.
Thrust-rate input MPC, Thrust input MPCを用いるそれぞれの場合において, 最小法線接触力を設定する場合と\SI{0}{N}にする4つのパターンで実験を行う.
}

\begin{figure}
    \centering
    \includegraphics[width=1.0\columnwidth]{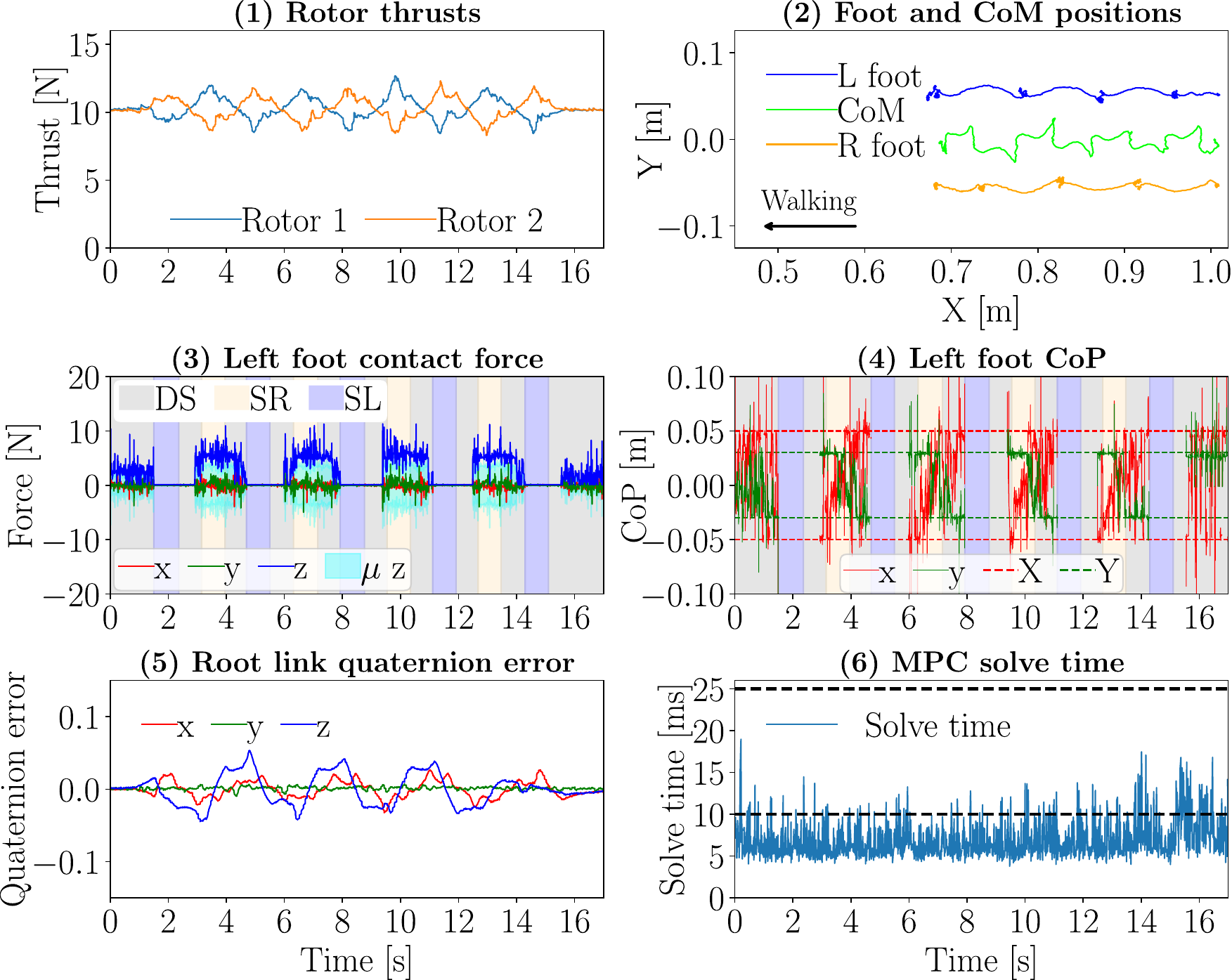}
    \caption{Results of anti-gravity walking simulation by thrust-rate input with \SI{5}{N} of minimum contact force. \revise{X and Y in (4) indicate the foot size in the \(x\) and \(y\) directions respectively.}}
    \vspace{-3mm}
    \label{figure:thrust_rate_minforce5}
\end{figure}


\subsubsection{Result}
\label{section:simulation_result}
\begin{figure}[t]
    \centering
    \includegraphics[width=1.0\columnwidth]{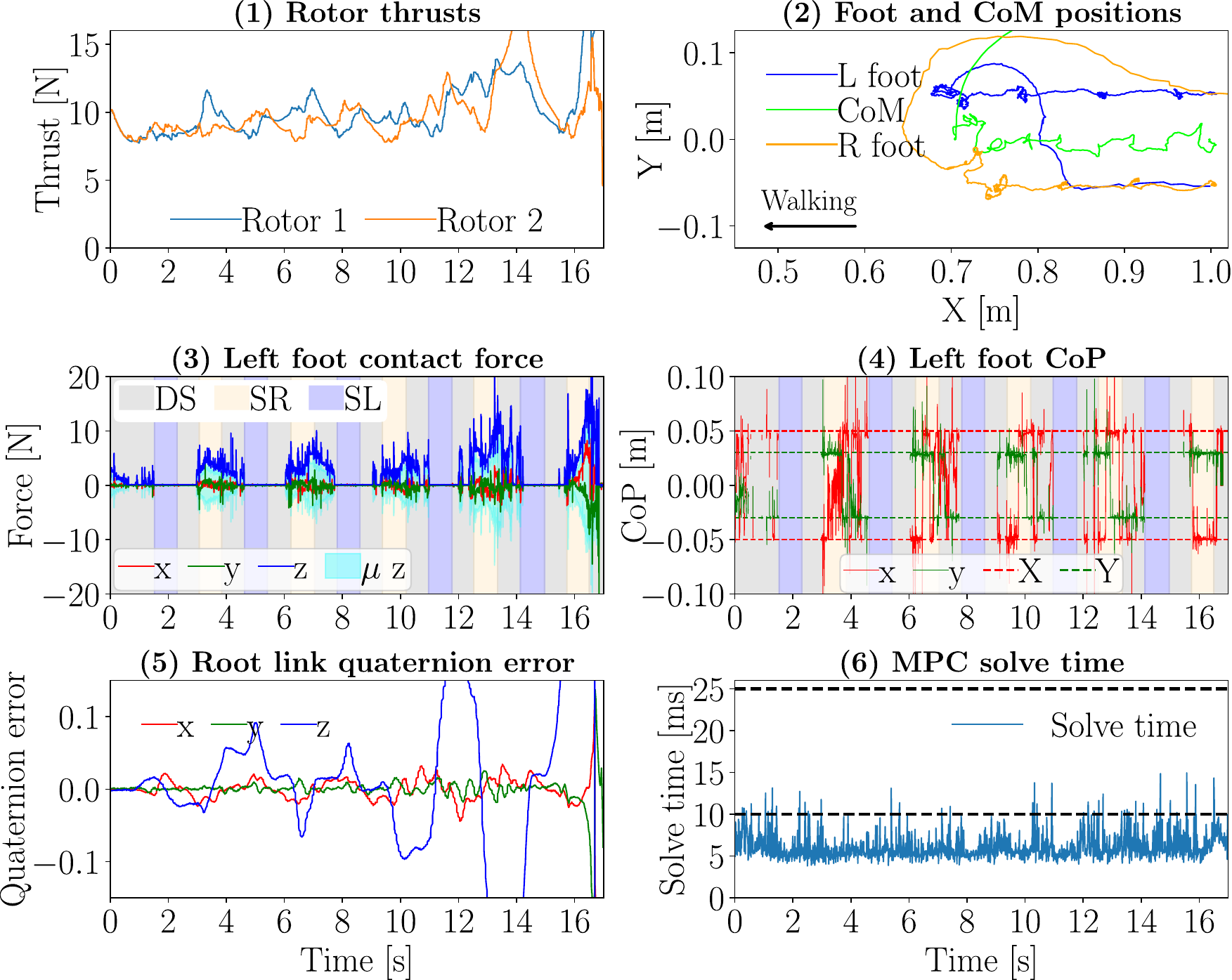}
    \caption{Results of anti-gravity walking simulation by thrust-rate input with \SI{0}{N} of minimum contact force.}
    \vspace{-3mm}
    \label{figure:thrust_rate_minforce0}
\end{figure}

\switchlanguage{
The anti-gravity walking motion achieved by using thrust-rate input MPC with \SI{5}{N} of minimum contact force \revise{and its results are} shown in \revise{the supplemental video} and \figref{figure:thrust_rate_minforce5}.
These plots show the \revise{commanded rotor} thrust, the projections of the feet and CoM onto the ceiling, the contact force of \revise{the left} foot obtained from MuJoCo sensor data, \revise{CoP position of the left foot}, the root (torso) link orientation error, and the MPC solv\revise{e} time.
The root link error is computed as the quaternion difference from \revise{its} target.

The generated thrust periodically and smoothly changed between about \SI{8}{N} and \SI{12}{N}.
In terms of the foot landing position, the deviation from the walking direction was kept below approximately \SI{0.02}{m}.
The foot contact force showed that during the swing phase, a normal force of \SI{5}{N} or more is generated at the support foot, maintaining stable contact.
Furthermore, during the double-support phase, the load assigned to last support foot was transferred to the next support foot and became zero for next swing phase.
\revise{Over the samples in which the left foot was in contact, i.e., the DS and SR phases in \figref{figure:thrust_rate_minforce5}(4), the CoP of the left foot deviated from its boundary in the \(x\)- and \(y\)-direction for \SI{5.00}{\%} and \SI{7.76}{\%} of the time, respectively, and the bound on the yaw torque was violated for \SI{3.01}{\%}.
Moreover, the duration of these violation was at most \SI{16}{ms}, and the system recovered within a few control cycles after the violation occurred.
The tangential force stayed well inside the friction cone: the friction utilization \(\sqrt{f^2_{{L_x}} + f^2_{L_y}}/\mu f_{L_z}\) at the left foot had a median of \num{0.24} and remained below \num{0.74} for \SI{95}{\%} of these samples.
While the CWC constraint was introduced as a penalty to prioritize real-time performance, the violation rate was kept low, and the system could recovered, so walking did not actually fail.}
The average MPC solving time was \SI{6.69}{ms}, \revise{and \SI{91.5}{\percent} of the solves completed within \SI{10}{ms}.}
\revise{Although the solve time occasionally exceeded \SI{10}{ms}, the worst case was \SI{18.9}{ms}, which is still shorter than the MPC discretization step \(\Delta t = \SI{25}{ms}\) (\tabref{table:robot_spec}).
Since the horizon is advanced according to the elapsed real time rather than each solve, as described in \secref{section:rti}, the solver kept up with the real-time MPC and the walking motion remained stable.}
}{
Thrust-rate MPCを用いて最小法線力を設定した場合に実現できた歩行動作を\figref{figure:overview}および添付動画に示す.
Thrust-rate MPCを用いて最小法線力を\SI{5}{N}, \SI{0}{N}にしたときの結果をそれぞれ\figref{figure:thrust_rate_minforce5}, \figref{figure:thrust_rate_minforce0}に示す.
これらはそれぞれ発揮推力, 足平と重心の天井への射影, 左足と右足の接触力, ルートリンクの姿勢誤差, MPCの求解時間を示している.
ルートリンク誤差は目標値とのクオータニオン差分をとったものである.

\figref{figure:overview}および添付動画に示すように, Thrust-rate MPCを用いて最小法線力を設定した場合には, ロボットは安定した抗重力歩行を実現している.
またこのとき発揮推力は最大で\SI{12}{N}程度, 最小で\SI{8}{N}程度を周期的に変化した.
また, 足の着地位置に着目すると進行方向からのズレは約\SI{0.02}{m}以下を実現することができた.
足平の接触力をみると, 遊脚期には支持脚側に\SI{5}{N}以上の法線方向の力を発生させることができ, 安定した接触を維持できていることがわかる.
さらに, 両足支持期には支持脚にかけていた体重を遷移させて遊脚期にゼロにすることができている.
\revise{CoP位置に着目すると, x方向, y方向について逸脱したのは全体の\SI{5.00}{\%}, \SI{7.76}{\%}となった.
また, 回転摩擦に関する違反率は, \SI{3.01}{\%}となった.
実時間性を優先し, CWC constraintはペナルティとして導入したが, 違反率は低く抑えられ, 違反後も数制御周期以内に回復し, 実際に歩行が破綻することはなかった.}
MPCの平均求解時間は\SI{6.69}{ms}であり, 全体の\SI{91.5}{\percent}を\SI{10}{ms}以内で解くことができた.
求解時間は稀に\SI{10}{ms}を超えたが, 最悪値でも\SI{18.9}{ms}であり, これはMPC離散化ステップ\(\Delta t = \SI{25}{ms}\)(\tabref{table:robot_spec})より短い.
\secref{section:rti}で述べたように, ホライズンは各求解ごとではなく経過実時間に応じて進めるため, 求解は実時間MPCに追従し, 歩行動作は安定に保たれた.
}

\subsubsection{\revise{Ablation study}}
\begin{figure}[t]
    \centering
    \includegraphics[width=1.0\columnwidth]{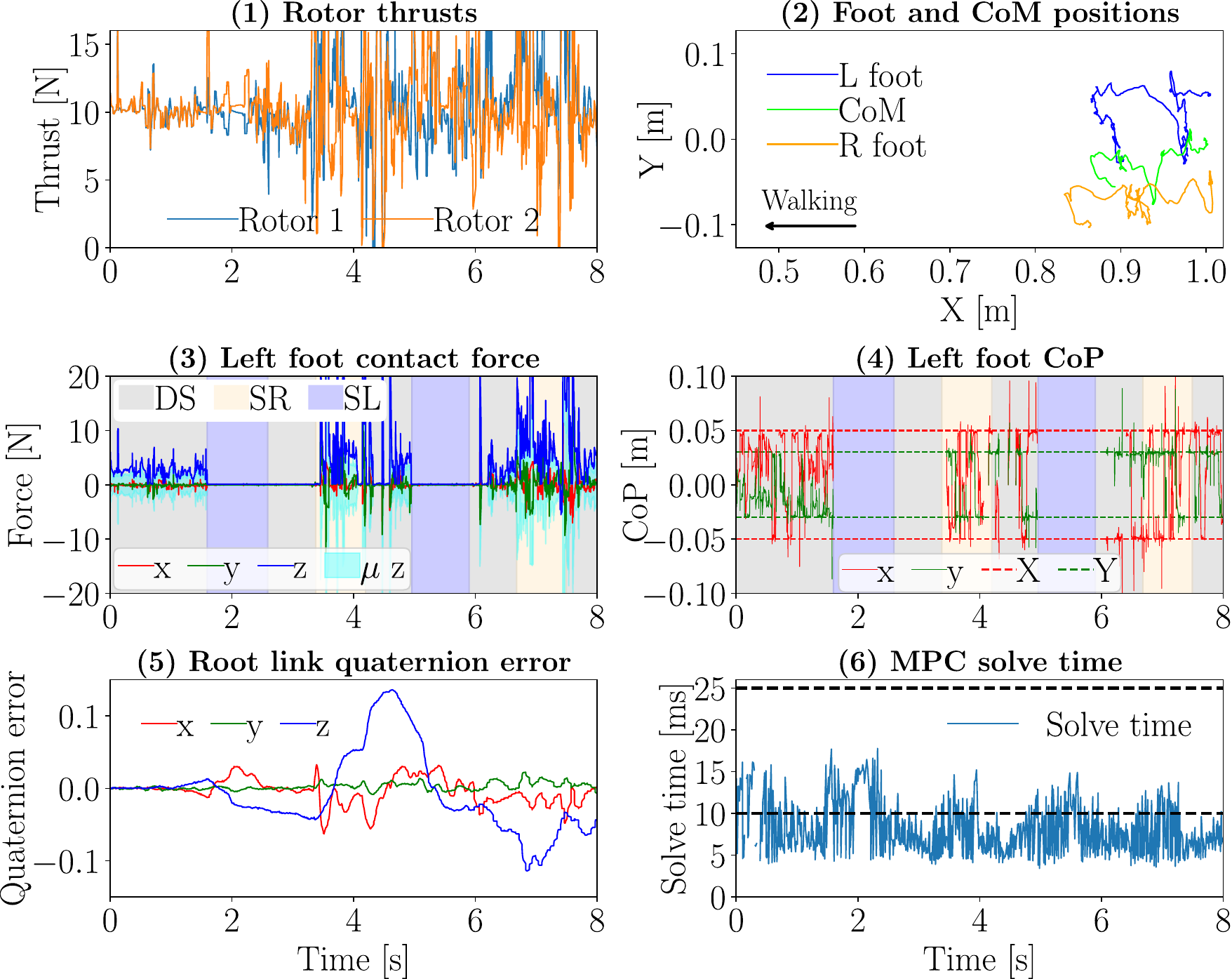}
    \caption{Results of anti-gravity walking simulation by thrust input with \SI{5}{N} of minimum contact force.}
    \vspace{-3mm}
    \label{figure:thrust_minforce5}
\end{figure}

\switchlanguage{
\revise{For comparison, we also conducted experiments without either of the proposed methods.}
\revise{\figref{figure:thrust_rate_minforce0} shows the results when the lower bounds of foot-normal component of contact force w\revise{ere} set to \SI{0}{N}}.
\revise{In this case}, the motion eventually became unstable and the robot detached from the ceiling.
\revise{I}n the first few steps, the generated thrust was approximately \SI{10}{N} for both thrusters.
These thrust magnitudes were smaller than \revise{in \secref{section:simulation_result}}.
However, the robot could not make stable contact with the ceiling especially in yaw direction.

Also, the result when using thrust input with \SI{5}{N} of minimum contact force is shown in \figref{figure:thrust_minforce5}.
In this case, thrust oscillated more significantly compared to the thrust-rate input case even in the initial stance phase.
After starting walking, \revise{this oscillation led to} unstable contact with the ceiling.
As a result, the entire body vibrated and eventually detached due to the foot leaving the ceiling.
At this time, the target thrust also reached the lower bound of \SI{0}{N}, indicating that physically plausible trajectories were not generated.

These results demonstrate that the proposed framework is effective for motions including contact switching in environments where natural support from gravity cannot be obtained.
}{
一方で, 足平法線方向に最小接触力を設定しなかった場合, 最終的に動作が不安定化し, 天井から遊離してしまった.
このケースでは最初の数歩における発揮推力は両方のスラスタで\SI{10}{N}程度となり, 最小接触力を設定する場合と比較して少ない発揮推力で動作ができていた.

また, 添付動画に示すように, Thrust-input MPCを利用した場合においてもシミュレーションを行った.
最小接触力を設定した場合においても, 初期の数歩は歩くことができたが, 目標推力が振動に応じて機体全体が振動し, 足裏が離れることで最終的には遊離してしまった.
このとき, 制御入力の境界である\SI{0}{N}を目標値とすることも発生しており, 物理的に妥当な軌道を生成できていないこともわかった.

以上の結果を通して, 本研究で提案したThrust-rateを入力とし, 足平法線方向に接触力の下限を設けた全身MPCが重力による自然な支持が得られない環境での接触の切り替えの伴う動作制御において有効であるということが示された.
}

\subsection{\revise{Hardware validation}}
\switchlanguage{\revise{
We conducted a hardware validation experiment.
The design parameters are shown in \figref{figure:dynamics}(a) and \tabref{table:robot_spec}.
For a lightweight design and sufficient joint torque, we deployed servo motors with a high reduction ratio, which were driven by position control.
Therefore, we only commanded the joint positions based on the optimal trajectories generated by the MPC.
Tethers to the power supply and control PC were attached; the effective mass, including other wiring, was \SI{2.0}{kg}.
Due to the communication delay between the control PC and the servos, we assumed during the swing phase that the joints follow the predicted trajectory, and used the joint positions and velocities of the previous optimal solution as the initial state.
To keep this open-loop integration from drifting, the foot and CoM references were corrected from measurement at every double-support phase.
Furthermore, since joint-level force control was difficult, thrust to stabilize the roll and pitch angles was added to the MPC outputs.
In this experiment, computation was performed with eight threads to keep a larger margin in the solve time.
}}{
\revise{
我々はハードウェアによる検証実験を行った.
設計パラメータは\figref{figure:dynamics}(a)および\tabref{table:robot_spec}に示す.
軽量設計と関節の十分な出力を両立すべく高減速比のサーボを搭載しており, 関節は位置制御で駆動される.
そのためMPCの最適軌道のうち関節位置を指令する.
地上から電源供給と制御PCのためのテザーを接続しており, その他配線のワイヤを含め実質的な質量は\SI{2.0}{kg}である.
通信速度の遅延による影響を抑えるべく, 遊脚期はMPCにおいて関節の状態フィードバックは利用せずオープンループで計画, 制御される.
また, 関節トルクによる力制御が難しいため, roll, pitch方向の姿勢の安定化については推力調整により補助的に行った.
また, 求解時間の余裕を確保するため, 実機では8スレッドで並列計算を行った.
}
}

\begin{figure}[t]
  \centering
  \includegraphics[width=1.0\columnwidth]{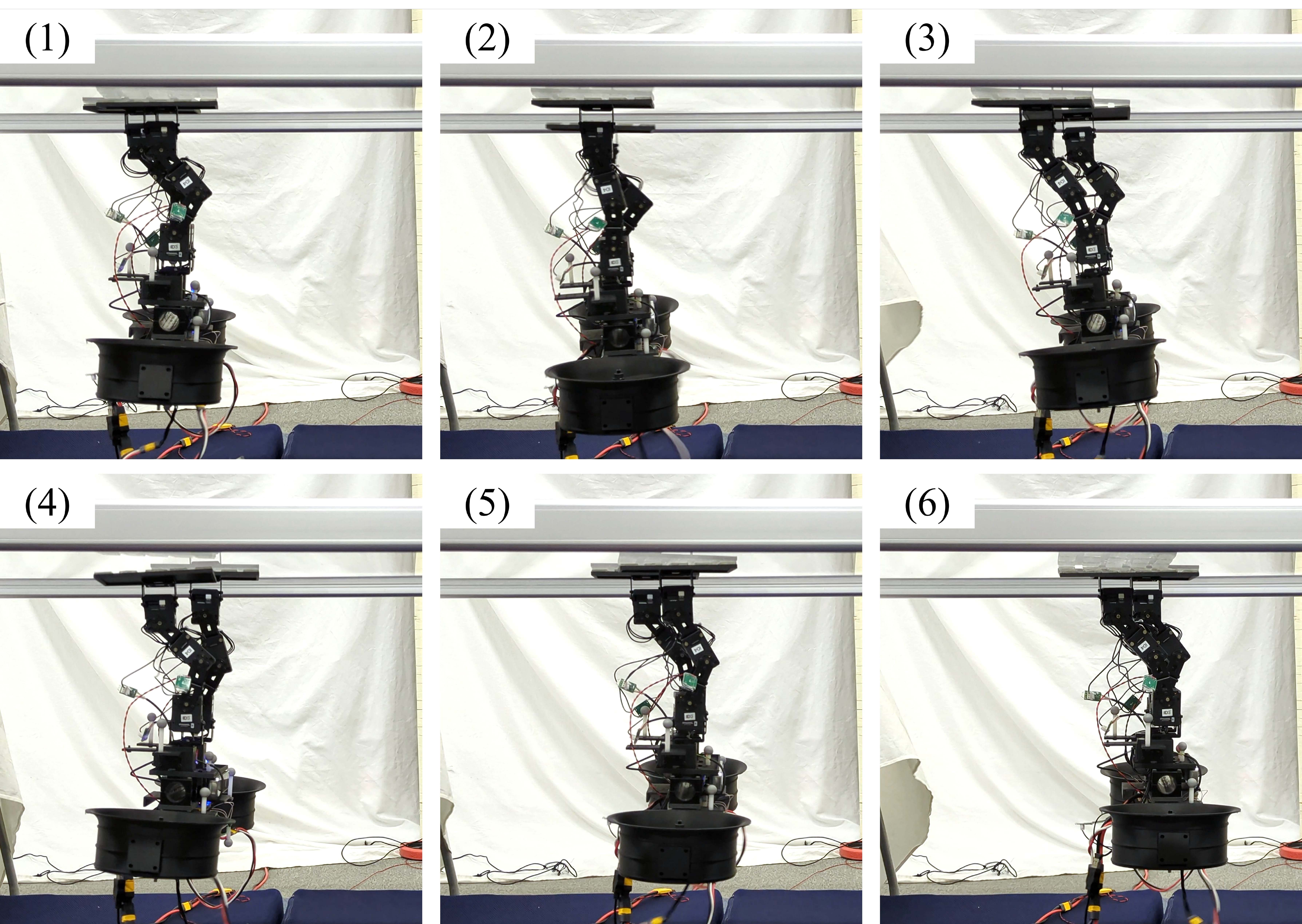}
    \caption{\revise{Hardware validation of anti-gravity walking motion.}}
    \vspace{0mm}
    \label{figure:mpc_real_machine}
\end{figure}

\begin{figure}[t]
  \centering
  \includegraphics[width=1.0\columnwidth]{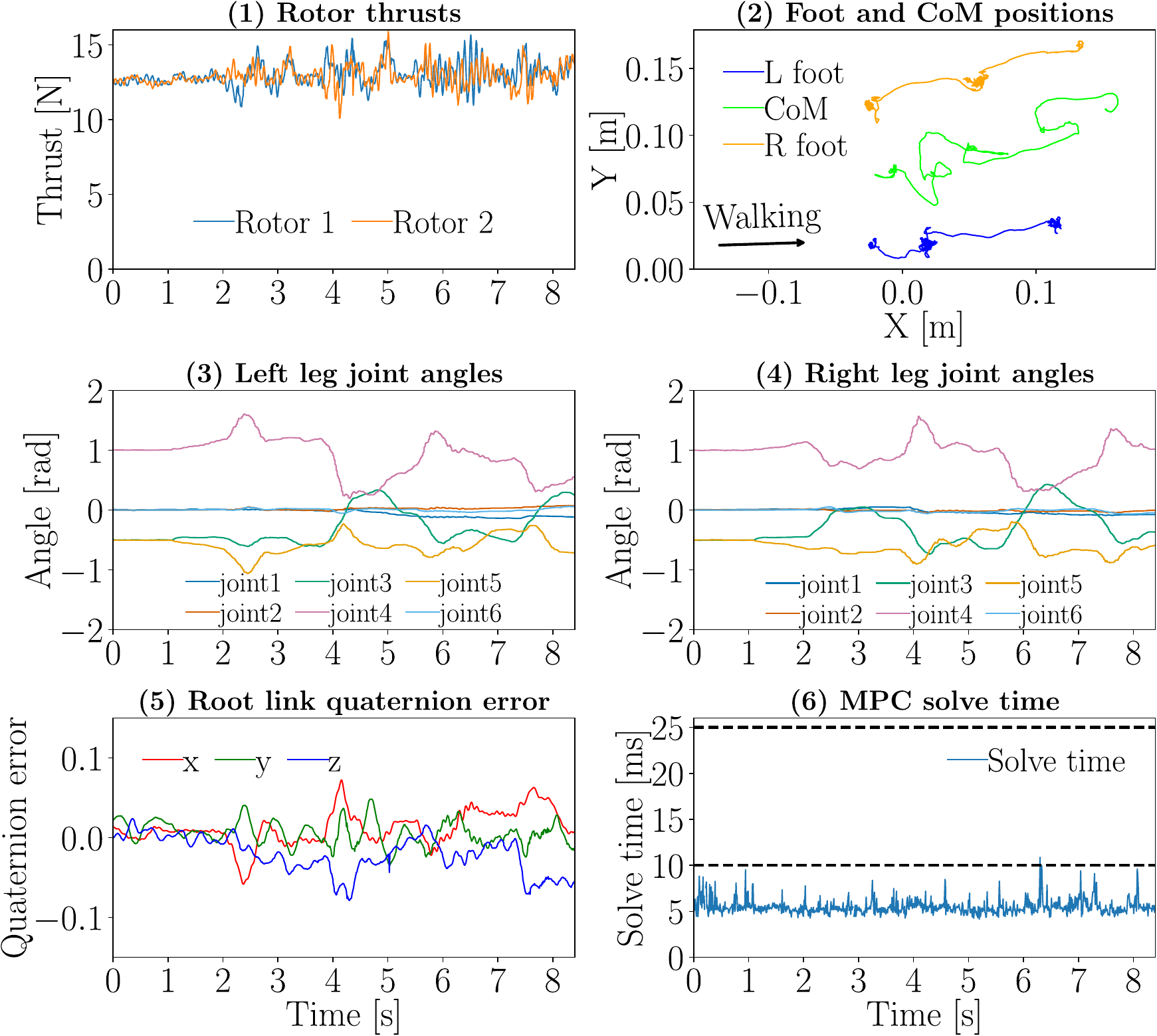}
  \caption{\revise{Plots related to \figref{figure:mpc_real_machine}.}}
  \vspace{-3mm}
  \label{figure:mpc_real_machine_data}
\end{figure}

\switchlanguage{\revise{
\figref{figure:mpc_real_machine} and supplemental video show the hardware experiment.
\figref{figure:mpc_real_machine_data} shows the commanded thrust, the projections of the feet and CoM onto the ceiling, the joint angles of each leg, the root link orientation error, and the MPC solve time.
The robot walked along the ceiling for approximately \SI{0.15}{m} in four steps, taking about \SI{8.5}{s}.
With the proposed formulation, thrust spikes and excessive vibrations were suppressed.
Since this robot was not equipped with wrench sensors on its feet, it was not possible to directly measure contact wrench or the CoP.
On the other hand, the thrust exerted by each thruster in steady state was approximately \SI{12.5}{N}, and its total exceeded the \SI{19.6}{N} corresponding to the robot's effective mass.
It is estimated that this difference between exerted thrust and gravity was used to generate contact force against the ceiling, and is comparable to the set \(F_{\min}\).
The orientation error was kept within \SI{0.1}{rad}.
The two rotors are separated along the \(y\)-axis, resulting in a long moment arm about the \(x\)-axis.
Consequently, the thrust on the left and right sides varies in response to the \(x\)-component of the orientation error; it can be observed that the larger the error, the greater the thrust of the left rotor (rotor1).
During this period, the average MPC solve time was \SI{5.47}{ms}, and \SI{99.7}{\percent} of the solutions were completed within \SI{10}{ms}.
}}{\revise{
\figref{figure:mpc_real_machine}にハードウェア実験の様子を示す.
また, この間の推力, 足と重心フレームの位置, 左右の脚の関節角度, ルートリンクの姿勢誤差, MPCの求解時間を\figref{figure:mpc_real_machine_data}に示す.
ロボットは4歩で\SI{8.5}{s}かけて\SI{0.15}{m}の天井歩行を実現した.
thrust-rateを入力とする定式化により推力スパイクや過剰な振動は抑えられた.
本機は足裏に力覚センサを搭載していないため, 接触レンチとCoPを直接計測することはできない.
一方で, 定常状態での各発揮推力は\SI{12.5}{N}程度であり, 2基の合計はロボットの実質的な質量に相当する\SI{19.6}{N}を上回る.
その差が足平を天井へ押し付ける接触力として生成されていると見積もられ, 設定した\(F_{\min}\)と同程度である.
姿勢誤差はいずれも\SI{0.1}{rad}に抑えられた.
左右のロータは進行方向である\(x\)軸に対して横方向に配置されており, \(x\)軸まわりのモーメントアームが長い.
そのため姿勢誤差の\(x\)成分に応じて左右の推力が変化し, 誤差が大きいほど左ロータ(rotor1)の推力が増加する様子が確認できる.
この間, MPCの平均求解時間は\SI{5.47}{ms}であり, 全体の\SI{99.7}{\percent}を\SI{10}{ms}以内で解くことができた.
}}

\switchlanguage{\revise{
Meanwhile, this experiment has the following limitations.
Because a high weight was assigned to swing foot and CoM tracking, the joint angles deviated from the initial state.
In particular, for the left knee (joint 4), the joint angle sometimes deviated from the initial \SI{1.0}{rad} to approximately \SI{0.3}{rad}.
Using joint angles generated by whole-body inverse kinematics as a reference for state regularization is expected to enable the robot to continue walking in a state that does not deviate from the initial state while still achieving the foot and CoM tracking tasks.
In this work, since we prioritized real-time computation and used a solver that handles only control input constraints, no equality constraints regarding the terminal state were imposed.
A solver capable of handling general constraints is expected to yield more stable motion.
Furthermore, due to the high reduction ratio of the servos, force control via joint torque was not possible, and posture stabilization was achieved using thrust.
By employing servos with a low reduction ratio, whole-body stabilization control integrating contact forces and thrust can be achieved.
}}{\revise{
一方で, 本実験には以下の限界がある.
足と重心の目標追従に高い重みを設定したため, 関節角度については初期姿勢から離れていき, 特に左膝(joint4)について, \SI{1.0}{rad}だったのが約\SI{0.3}{rad}へと伸びたような姿勢になることがあった.
全身逆運動学で生成した関節角度を状態正則化の参照に用いることで, 足先と重心の追従タスクを実現しつつ初期姿勢から逸脱しない姿勢で歩行を続けられると考えられる.
また, 今回は実時間での求解を優先し制御入力の制約のみを扱うソルバーを用いたため, 終端姿勢に関する等式制約などは課せなった.
一般の制約を扱えるソルバー\cite{hpipm}を用いればより安定な動作が期待できる.
さらに,  高減速比サーボに起因して関節トルクによる力制御が行えず, 姿勢安定化は推力のみで行った.
低減速比サーボを用いることで, 接触力と推力を複合した全身の安定化制御が実現できると考えられる.
}
}
\section{Conclusion}
\label{sec:conclusion}
\switchlanguage{
In this paper, we propose a thrust-rate input whole-body MPC for anti-gravity walking by a flying humanoid.
We augmented the state with thrust and formulated the problem with thrust-rate as the control input, which maintains thrust continuity even at contact switching.
Compared with a conventional \revise{thrust input formulation}, the thrust-rate input formulation suppressed \revise{spikes in} thrust, contact forces, and joint torques, and the convergence of trajectory optimization became faster.
In addition, we propose a load transfer strategy that treats the minimum foot-normal component of contact force  as a lower bound of the CWC \revise{penalty} and smoothly transfers them during the double-support phase.
We integrated these components into a real-time control framework and demonstrated anti-gravity walking in dynamics simulation \revise{and validated on hardware}.

\revise{In the future,} applying the \revise{proposed framework} to more general three-dimensional environments, such as transitions from walls to ceilings and walking on arbitrary curved surfaces, and integrating it with \revise{aerial flight} would further expand the applicability of flying humanoid robots.
}{
本論文では, 飛行ヒューマノイドによる抗重力歩行を実現するための推力差分入力型全身MPCを提案した.
状態に推力を加える拡張をし, 推力の時間微分を制御入力とすることで, 接触切替え時にも推力軌道の連続性を保つ定式化を導入した.
Thrust-rate Input MPCは推力を入力として扱うナイーブな定式化と比較して, 接触切替え時の推力スパイクを抑制できた.
推力スパイクの抑制により, 足平での接触力や関節トルクも滑らかになり, 軌道最適化の収束も速くなることがわかった.
また, 歩行安定化のために必要な足裏の最小押し付け力を接触レンチ錐制約の下限として扱い, 両脚支持期に左右足間でこの最小法線力を滑らかに遷移させる荷重移動戦略を提案した.
そして, これらを統合した実時間制御フレームワークを構築し, シミュレーションで抗重力歩行動作が可能なことを実証した.

今後の課題として, 実機における推力応答遅れを考慮して状態フィードバックにいれることや, 接触面の摩擦・形状誤差, 足裏接触検出の不確かさを含めたロバスト化が挙げられる.
また, 壁面付近での流体力学的特性の考慮することでより安定な動作になると考えられる.
さらに, 壁面から天井面への遷移や任意曲面上での歩行など, より一般的な三次元環境への適用や飛行動作との統合をすることで飛行ヒューマノイドの応用可能性を広げられるだろう.
}

\bibliographystyle{junsrt}
\bibliography{main}

\begin{thebibliography}{10}

\bibitem{kim2021bipedal}
Kyunam Kim, et~al.
\newblock A bipedal walking robot that can fly, slackline, and skateboard.
\newblock {\em Science Robotics}, Vol.~6, No.~59, p. eabf8136, 2021.

\bibitem{anzai2021design}
Tomoki Anzai, et~al.
\newblock Design and development of a flying humanoid robot platform with
  bi-copter flight unit.
\newblock In {\em 2020 IEEE-RAS 20th International Conference on Humanoid
  Robots}, pp. 69--75. IEEE, 2021.

\bibitem{gorbani2025ironcub}
Davide Gorbani, et~al.
\newblock ironcub 3: The jet-powered flying humanoid robot.
\newblock {\em arXiv preprint arXiv:2506.01125}, 2025.

\bibitem{xu2025system}
Bo~Xu, et~al.
\newblock System design and flight control of a flying wheel-legged humanoid
  robot (fwlr).
\newblock {\em Industrial Robot: the international journal of robotics research
  and application}, 2025.

\bibitem{sugihara2023design}
Kazuki Sugihara, et~al.
\newblock Design and control of a small humanoid equipped with flight unit and
  wheels for multimodal locomotion.
\newblock {\em IEEE Robotics and Automation Letters}, Vol.~8, No.~9, pp.
  5608--5615, 2023.

\bibitem{khazoom2024tailoring}
Charles Khazoom, et~al.
\newblock Tailoring solution accuracy for fast whole-body model predictive
  control of legged robots.
\newblock {\em IEEE Robotics and Automation Letters}, Vol.~9, No.~12, pp.
  11074--11081, 2024.

\bibitem{boxfddp}
Carlos Mastalli, et~al.
\newblock A feasibility-driven approach to control-limited ddp.
\newblock {\em Autonomous Robots}, Vol.~46, No.~8, pp. 985--1005, 2022.

\bibitem{zhao2023spidar}
Moju Zhao, et~al.
\newblock Design, modeling, and control of a quadruped robot spidar:
  Spherically vectorable and distributed rotors assisted air-ground quadruped
  robot.
\newblock {\em IEEE Robotics and Automation Letters}, Vol.~8, No.~7, pp.
  3923--3930, 2023.

\bibitem{wang2025dynamic}
Chenghao Wang, et~al.
\newblock Dynamic quadrupedal legged and aerial locomotion via structure
  repurposing.
\newblock In {\em 2025 IEEE/RSJ International Conference on Intelligent Robots
  and Systems}, pp. 10664--10669. IEEE, 2025.

\bibitem{liu2018jet}
Biao Liu, et~al.
\newblock Jet-hr1: Stepping posture optimization for bipedal robot over large
  ditch based on a ducted-fan propulsion system.
\newblock In {\em 2018 IEEE/RSJ International Conference on Intelligent Robots
  and Systems}, pp. 6010--6015. IEEE, 2018.

\bibitem{dangol2021control}
Pravin Dangol, et~al.
\newblock Control of thruster-assisted, bipedal legged locomotion of the harpy
  robot.
\newblock {\em Frontiers in Robotics and AI}, Vol.~8, p. 770514, 2021.

\bibitem{salagame2024quadrupedal}
Adarsh Salagame, et~al.
\newblock Quadrupedal locomotion control on inclined surfaces using collocation
  method.
\newblock In {\em 2024 American Control Conference}, pp. 2838--2843. IEEE,
  2024.

\bibitem{zhang2025flight}
Yan Zhang, et~al.
\newblock Flight poses optimization and hierarchical control strategy for
  flying humanoid robots.
\newblock {\em Advanced Robotics}, Vol.~39, No.~22, pp. 1395--1417, 2025.

\bibitem{ollero2022aerialmanipulationreview}
Anibal Ollero, et~al.
\newblock Past, present, and future of aerial robotic manipulators.
\newblock {\em IEEE Transactions on Robotics}, Vol.~38, No.~1, pp. 626--645,
  2022.

\bibitem{bodie2019omnidirectionalmanipulation}
Karen Bodie, et~al.
\newblock {An Omnidirectional Aerial Manipulation Platform for Contact-Based
  Inspection}.
\newblock In {\em Proceedings of Robotics: Science and Systems},
  FreiburgimBreisgau, Germany, June 2019.

\bibitem{nishio2023perchingarm}
Takuzumi Nishio, et~al.
\newblock Design, control, and motion planning for a root-perching
  rotor-distributed manipulator.
\newblock {\em IEEE Transactions on Robotics}, Vol.~40, pp. 660--676, 2023.

\bibitem{kim2025contactimplicit}
Gijeong Kim, et~al.
\newblock Contact-implicit model predictive control: Controlling diverse
  quadruped motions without pre-planned contact modes or trajectories.
\newblock {\em The International Journal of Robotics Research}, Vol.~44, No.~3,
  pp. 486--510, 2025.

\bibitem{radosavovic2024real}
Ilija Radosavovic, et~al.
\newblock Real-world humanoid locomotion with reinforcement learning.
\newblock {\em Science Robotics}, Vol.~9, No.~89, p. eadi9579, 2024.

\bibitem{shi2019dynamics}
Fan Shi, et~al.
\newblock Multi-rigid-body dynamics and online model predictive control for
  transformable multi-links aerial robot.
\newblock {\em Advanced Robotics}, Vol.~33, No.~19, pp. 971--984, 2019.

\bibitem{martisaumell2023fullbodytorquelevelnonlinearmodel}
Mart{\'\i}-Saumell, et~al.
\newblock Full-body torque-level non-linear model predictive control for aerial
  manipulation.
\newblock {\em arXiv preprint arXiv:2107.03722}, 2021.

\bibitem{brunner2020trajectorytracking}
Maximilian Brunner, et~al.
\newblock Trajectory tracking nonlinear model predictive control for an
  overactuated mav.
\newblock In {\em 2020 IEEE International Conference on Robotics and
  Automation}, pp. 5342--5348, 2020.

\bibitem{mastalli2020crocoddyl}
Carlos Mastalli, et~al.
\newblock Crocoddyl: An efficient and versatile framework for multi-contact
  optimal control.
\newblock In {\em 2020 IEEE International Conference on Robotics and
  Automation}, pp. 2536--2542. IEEE, 2020.

\bibitem{hpipm}
Gianluca Frison, et~al.
\newblock Hpipm: a high-performance quadratic programming framework for model
  predictive control.
\newblock {\em IFAC-PapersOnLine}, Vol.~53, No.~2, pp. 6563--6569, 2020.

\bibitem{caron2015stability}
St{\'e}phane Caron, et~al.
\newblock Stability of surface contacts for humanoid robots: Closed-form
  formulae of the contact wrench cone for rectangular support areas.
\newblock In {\em 2015 IEEE International Conference on Robotics and
  Automation}, pp. 5107--5112. IEEE, 2015.

\bibitem{todorov2012mujoco}
Emanuel Todorov, et~al.
\newblock Mujoco: A physics engine for model-based control.
\newblock In {\em 2012 IEEE/RSJ international conference on intelligent robots
  and systems}, pp. 5026--5033. IEEE, 2012.

\end{thebibliography}

\end{document}